\documentclass[11pt,a4paper,logo,onecolumn]{trajlong_report}
\usepackage[authoryear,sort&compress,round]{natbib}
\usepackage{microtype}
\definecolor{myCite}{HTML}{1C4587}
\hypersetup{citecolor=myCite,linkcolor=myCite,urlcolor=myCite}

\IfFileExists{OT1XCharter-TLF.fd}{%
  \SetSymbolFont{operators}{normal}{OT1}{XCharter-TLF}{m}{n}
  \SetSymbolFont{operators}{bold}{OT1}{XCharter-TLF}{b}{n}
  \SetMathAlphabet{\mathit}{normal}{OT1}{XCharter-TLF}{m}{it}
  \SetMathAlphabet{\mathit}{bold}{OT1}{XCharter-TLF}{b}{it}
  \SetMathAlphabet{\mathbf}{normal}{OT1}{XCharter-TLF}{b}{n}
  \SetMathAlphabet{\mathbf}{bold}{OT1}{XCharter-TLF}{b}{n}
}{}

\usepackage{amsmath,amsfonts,bm}

\def\eqref#1{equation~\ref{#1}}

\def\1{\bm{1}}

\DeclareMathAlphabet{\mathsfit}{\encodingdefault}{\sfdefault}{m}{sl}
\SetMathAlphabet{\mathsfit}{bold}{\encodingdefault}{\sfdefault}{bx}{n}

\usepackage{hyperref}
\usepackage{url}

\usepackage{graphicx}
\usepackage{placeins} 
\usepackage{booktabs}
\usepackage{multirow}

\newif\ifshowdraftnotes
\showdraftnotestrue 
\DeclareRobustCommand{\draftnote}[1]{\ifshowdraftnotes\textcolor{red}{#1}\fi}
\usepackage{amsmath}
\usepackage{amssymb}
\usepackage{mathtools}
\usepackage{amsthm}
\usepackage{pifont}
\usepackage{tcolorbox} 
\tcbuselibrary{breakable} 
\usepackage{soul} 
\usepackage{tikz} 
\usepackage{colortbl} 
\usepackage{adjustbox} 
\usepackage{wrapfig} 
\usepackage{needspace} 
\usepackage{enumitem} 
\usepackage{subcaption} 
\definecolor{darkgreen}{rgb}{0.0, 0.5, 0.0} 
\definecolor{darkblue}{RGB}{30, 100, 180} 
\definecolor{softpink}{RGB}{220, 100, 130} 
\definecolor{TrajRetrieval}{HTML}{AD6836}
\definecolor{TrajMultidocQA}{HTML}{885687}
\definecolor{TrajStateTracking}{HTML}{367B9D}
\definecolor{TrajAll}{HTML}{B59836} 

\usepackage{fontspec}
\usepackage{xeCJK}
\setCJKsansfont{FandolHei-Regular.otf}[BoldFont=FandolHei-Bold.otf]
\setCJKmonofont{FandolFang-Regular.otf}
\setCJKfamilyfont{CaseJP}{HaranoAjiMincho-Regular.otf}
\newfontfamily\CaseSymbolFont{DejaVuSans}[Extension=.ttf]
\usepackage{longtable}
\usepackage{array}

\definecolor{CaseAnswer}{HTML}{397666}

\AtBeginEnvironment{longtable}{\addfontfeatures{Mapping=}}

\definecolor{EffectGreen}{cmyk}{0.91,0,0.88,0.12}
\definecolor{EffectRed}{cmyk}{0,0.89,0.94,0.28}

\newcommand{\posdiff}[1]{%
  \ensuremath{\textcolor{EffectGreen}{_{\scriptstyle +#1}}}}
\newcommand{\negdiff}[1]{%
  \ensuremath{\textcolor{EffectRed}{_{\scriptstyle -#1}}}}
\newcommand{\zerodiff}{%
  \ensuremath{\textcolor{gray}{_{\scriptstyle +0.0}}}}

\newcommand{\OurMODEL}{\textsc{TrajLong}}

\title{\OurMODEL{}: Co-Designing Agentic and Long-Context Supervision for Mid-Training}

\renewcommand{\reportlogos}{%
  \includegraphics[height=17.05pt]{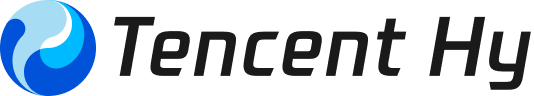}%
  \hspace{12pt}%
  \includegraphics[height=17.05pt]{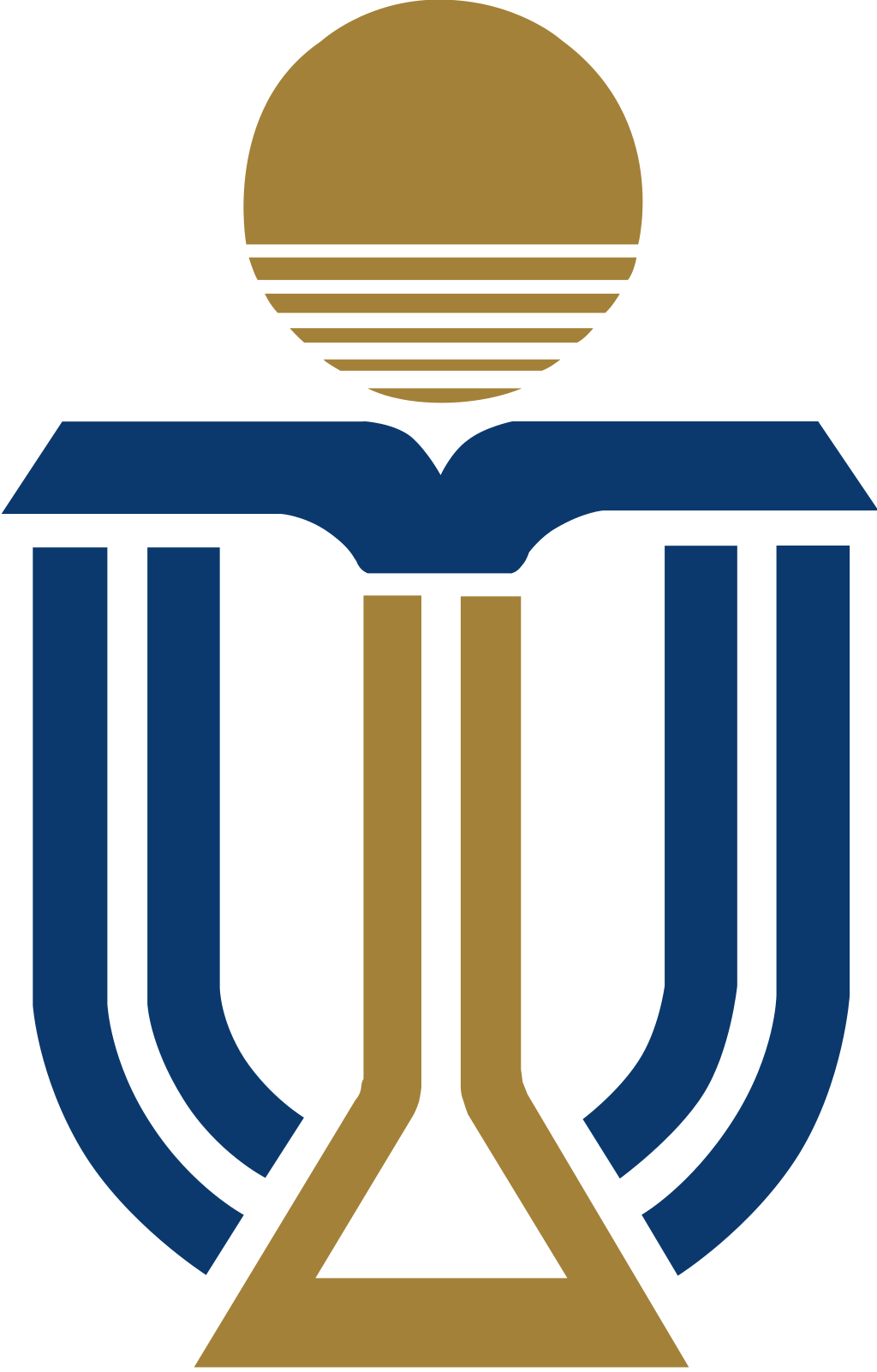}%
  \hspace{12pt}%
  \includegraphics[height=17.05pt]{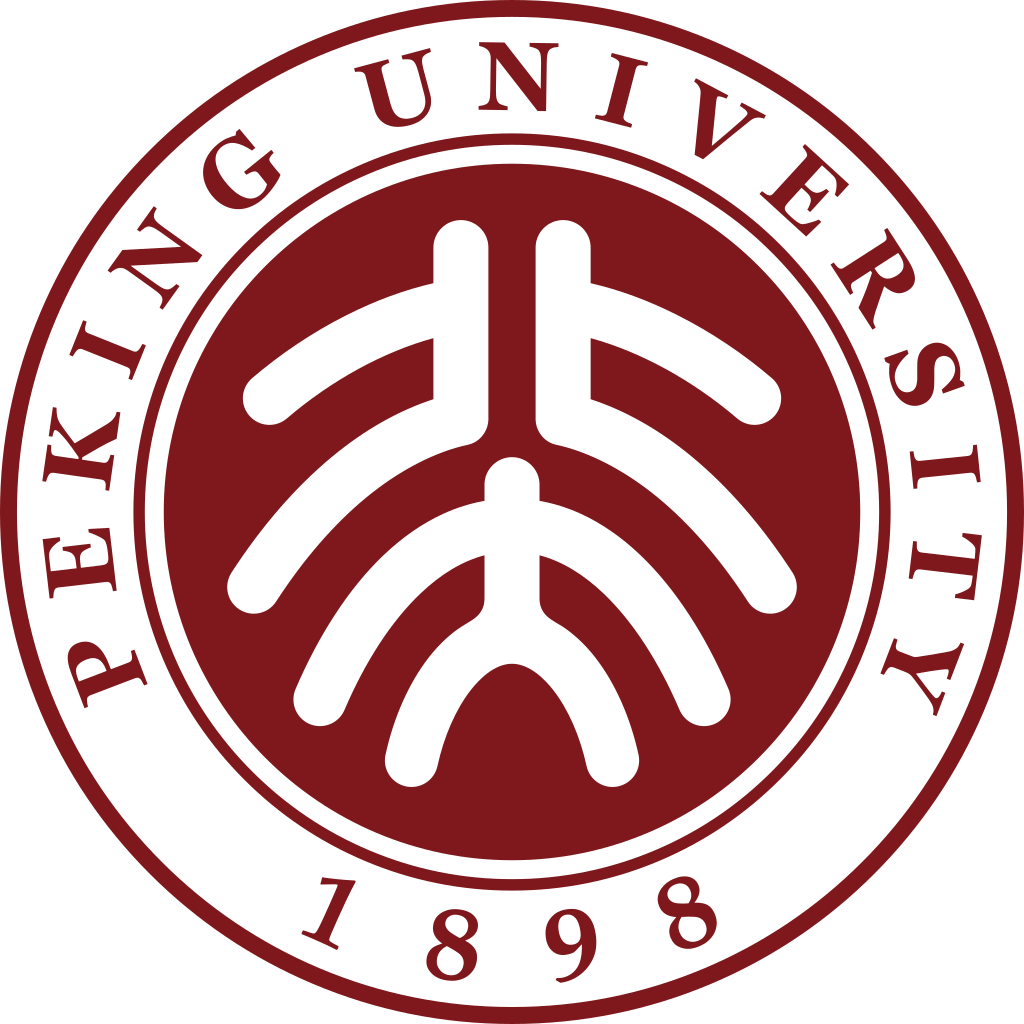}%
}

\DeclareRobustCommand{\trajlongauthorblock}{%
  \fontsize{10.5}{14}\selectfont
  Miao Peng\textsuperscript{1,2*},
  Qintong Zhang\textsuperscript{2,3*},
  Nuo Chen\textsuperscript{2\textdagger\textdaggerdbl},
  Yuhan Li\textsuperscript{2},
  Guochen Yan\textsuperscript{2,3}\\
  Xinran Gu\textsuperscript{2},
  Hongqiu Wu\textsuperscript{2},
  Hai Wang\textsuperscript{2},
  Lydell Huang\textsuperscript{2},
  Wentao Zhang\textsuperscript{3\textdagger},
  Jia Li\textsuperscript{1\textdagger}\\[4pt]
  \normalfont\fontsize{9.5}{12}\selectfont
  \textsuperscript{1}The Hong Kong University of Science and Technology (Guangzhou)\\
  \textsuperscript{2}Hunyuan Team, Tencent\qquad
  \textsuperscript{3}Peking University\\[3pt]
  {\fontsize{8.5}{11}\selectfont\ttfamily mpeng885@connect.hkust-gz.edu.cn\quad chennuo26@gmail.com}\\
  {\fontsize{8.5}{11}\selectfont\ttfamily wentao.zhang@pku.edu.cn\quad jialee@hkust-gz.edu.cn}%
}
\author{\trajlongauthorblock}
\AddToHook{cmd/maketitle/after}{%
  \begingroup
  \footnotetext[1]{Equal contribution.}%
  \footnotetext[3]{Project Lead.}%
  \footnotetext[2]{Corresponding authors.}%
  \endgroup
}
\hypersetup{pdfauthor={Miao Peng, Qintong Zhang, Nuo Chen, Yuhan Li, Guochen Yan, Xinran Gu, Hongqiu Wu, Hai Wang, Lydell Huang, Wentao Zhang, Jia Li}}

\begin{abstract}
  LLM agents for coding, search, and workplace tasks increasingly rely on long-context capabilities to effectively aggregate and reason over extended interaction histories. Recent work has incorporated agent trajectories into mid-training stage, drawing on their naturally long and interaction-rich structure. Yet how to organize the information within these trajectories into effective mid-training supervision remains underexplored. In this work, we investigate the relationship between long-context and agent atomic capabilities and introduce \textbf{\OurMODEL{}}, a novel framework that compiles trajectories into long-context training tasks with dense supervision, targeting three representative atomic capabilities: evidence grounding, cross-evidence aggregation, and temporal state maintenance. We mid-train Qwen3-14B-Base and Qwen3-30B-A3B-Base with data compiled by \OurMODEL{}, followed by supervised fine-tuning. Experiments on six long-context and twelve agent benchmarks demonstrate broad performance gains, with controlled ablations showing improvements over raw and masked trajectory baselines. Capability-level analyses further reveal task-dependent associations between long-context and agent atomic capabilities. These findings suggest that the shared capability demands of long-context reasoning and agent execution provide a principled basis for designing mid-training data to develop downstream agent capabilities.
\end{abstract}

\begin{document}
\maketitle

\section{Introduction}
\label{sec:introduction}

Large language models (LLMs) are increasingly expected to serve as agents in a wide range of real-world applications including coding, search, and workplace tasks~\citep{DBLP:journals/corr/abs-2406-12045, ToolACE, Dr_Tulu}. Completing these long-horizon user tasks requires LLMs to integrate information across many interaction turns, making effective use of long context essential to agent execution~\citep{SWE_bench,zhou2026deep,su2026acc}. Agent trajectories capture these extended interaction histories and provide a natural source of agent-oriented supervision. Recent work incorporates such supervision into mid-training, exposing base models to tool-use patterns and evidence-guided navigation before subsequent adaptation~\citep{jiang2026midtool, Scaling_CPT}. Beyond teaching interaction patterns, these trajectories also offer rich long-context training material for developing the foundational capabilities required for agent execution.

\begin{figure}[!t]
    \centering
    \begin{minipage}[c]{0.48\linewidth}
        \centering
        \includegraphics[width=\linewidth]{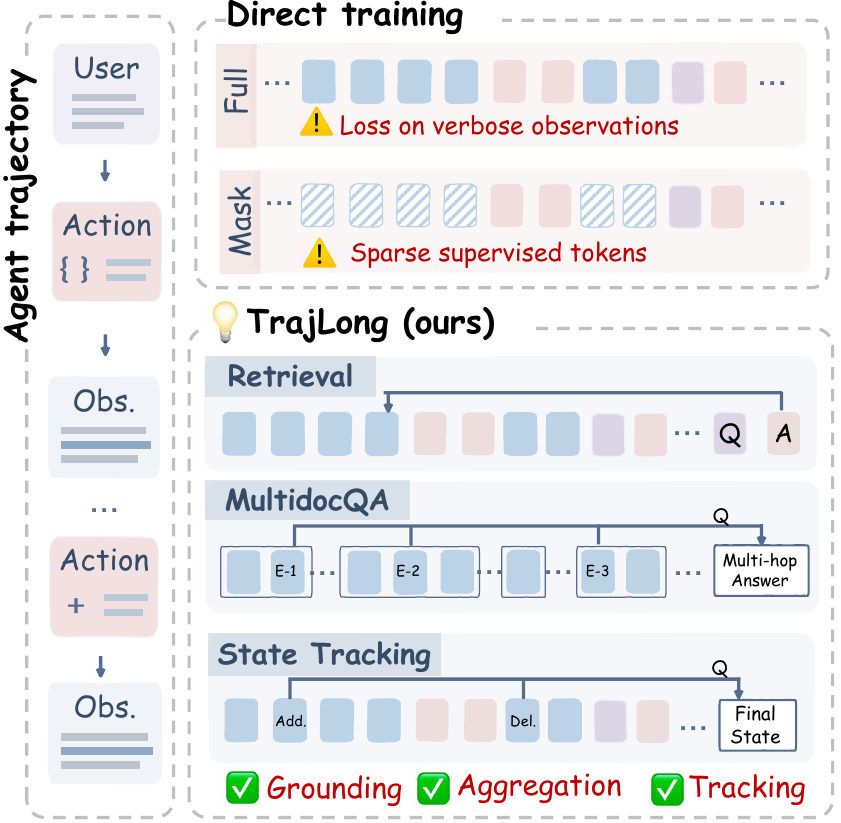}
    \end{minipage}\hspace{0.027\linewidth}%
    \begin{minipage}[c]{0.44\linewidth}
        \centering
        \includegraphics[width=\linewidth]{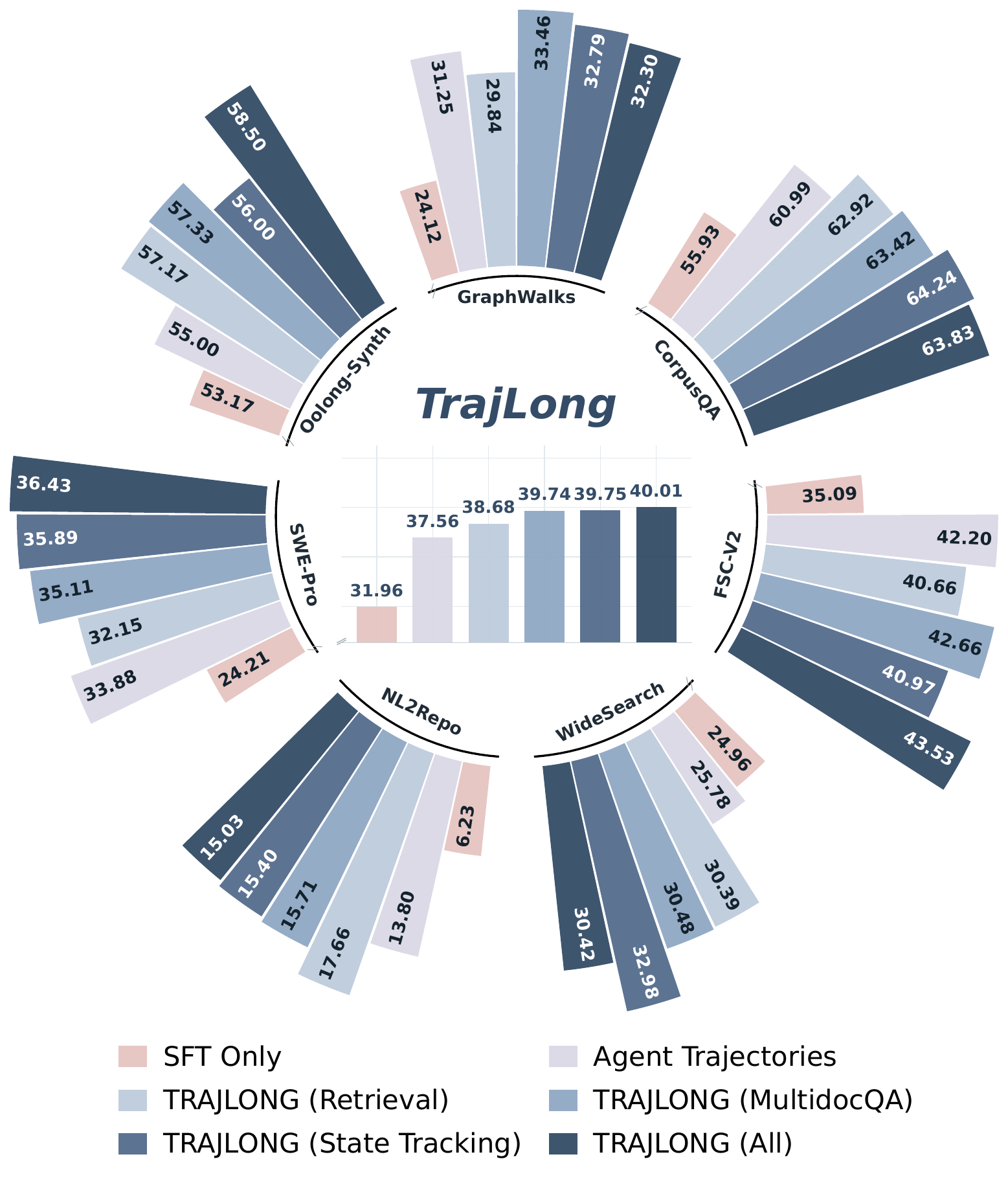}
    \end{minipage}
    \caption{Overview of \OurMODEL{}. Left: Full-trajectory and masked mid-training strategies compared with \OurMODEL{} compilation. Right: Avg@3 scores of Qwen3-14B variants on seven long-context and agent benchmarks across search, tool-use, and coding tasks.}
    \label{fig:intro_overview}
\end{figure}

The challenge is how to turn this rich but heterogeneous trajectory information into effective mid-training supervision. Applying next-token prediction to complete trajectories computes loss on both agent outputs and tool observations, introducing noisy tool responses into the training targets~\citep{daVinci_Dev, SWE_Universe, Kimi_Dev}. To control these training signals, a common strategy is loss masking, which excludes selected tokens from the training loss. For example, masking observations and other non-assistant content concentrates supervision on agent reasoning and actions~\citep{Youtu_LLM, zhou2026deep}. However, masked content must still be processed as context, reducing training token efficiency. Figure~\ref{fig:intro_overview} (left) contrasts full-trajectory and masked supervision. These choices determine where loss is applied, but do not by themselves organize long-range information dependencies into targeted learning tasks. ACC moves beyond loss allocation by compiling trajectory evidence into long-context question--answer examples for supervised fine-tuning, while retaining the original task question~\citep{su2026acc}. How to organize this information to systematically target distinct long-context capabilities remains open.

\begingroup
\setlength{\columnsep}{12pt}
\setlength{\intextsep}{6pt}
\begin{wrapfigure}{R}{0.44\textwidth}
    \centering
    \includegraphics[width=0.96\linewidth]{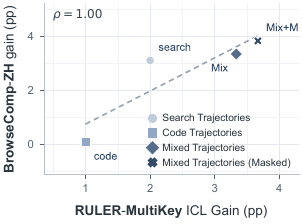}
    \captionsetup{font=small,justification=raggedright,singlelinecheck=false,skip=3pt}
    \caption{Correlation between RULER ICL gains and downstream BrowseComp-ZH gains across Qwen3-14B variants mid-trained with different agent trajectories.}
    \label{fig:icl_to_sft_retrieval_intro}
    \vspace{-\baselineskip} %
\end{wrapfigure}

The information dependencies within these trajectories provide a concrete basis for designing such supervision. Similar search results require distinguishing a specific fact from related distractors; complementary observations must be combined across distant turns; and successive edits can invalidate earlier information about the same object. These dependencies resemble the retrieval, multi-hop reasoning, and aggregation demands of long-context evaluation~\citep{hsieh2024ruler}, while adding the need to reconstruct evolving state~\citep{LongMemEval,EvoMemBench}. This suggests that long-context reasoning and agentic execution may share capability primitives, providing a basis for organizing trajectory data around the information processing required by downstream real-world tasks.

Figure~\ref{fig:icl_to_sft_retrieval_intro} offers an exploratory view of this connection. Across four training settings, RULER-MultiKey ICL gains correlate positively with BrowseComp-ZH gains after the same SFT recipe ($\rho=1.00$). It suggests a possible correlation between long-context capability and downstream agentic performance. This further motivates moving beyond aggregate long-context scores to examine which long-context capabilities are relevant to different agent atomic abilities and how they can guide the organization of trajectory data. We therefore ask: \textbf{\emph{How can the information dependencies within agent trajectories be organized into effective long-context supervision for agentic mid-training?}}

We introduce \textbf{\OurMODEL{}}, a capability-targeted framework for agentic mid-training (Figure~\ref{fig:intro_overview}). It reorganizes agent trajectory context and constructs questions and target responses around the underlying information dependencies, targeting evidence grounding (Retrieval), cross-evidence aggregation (MultidocQA), and temporal state maintenance (State Tracking). We apply next-token prediction across all tokens in the compiled context, question, and target response.

We perform capability-targeted mid-training on Qwen3-14B-Base and Qwen3-30B-A3B-Base, followed by SFT, achieving gains over SFT-only baselines across diverse long-context benchmarks and search, tool-use, and coding agent benchmarks. Figure~\ref{fig:intro_overview} (right) summarizes post-SFT performance of the Qwen3-14B variants on seven selected benchmarks. Matched-budget Qwen3-14B ablations further show that the combined variant outperforms both unmasked and masked raw-trajectory mixtures. Our analyses during mid-training and after SFT reveal overlapping long-context and agent gains with task-dependent learning dynamics. Evidence-based QA and dependency reasoning are associated with code understanding and planning, while state-oriented supervision may also support aggregation and structured reasoning beyond temporal tasks, suggesting shared information-processing demands between long-context tasks and agent execution.

Our contributions are summarized as follows: \textbf{(1)} We introduce \OurMODEL{}, a trajectory compilation framework targeting evidence grounding, cross-evidence aggregation, and temporal state maintenance for long-context agentic mid-training. \textbf{(2)} We demonstrate broad downstream improvements on both 14B and 30B models across long-context reasoning, search, tool use, and coding agent tasks. \textbf{(3)} We characterize task-dependent correlations between long-context performance and agent atomic capabilities, and examine the utility of compiled supervision across mid-training and SFT.
\par
\endgroup

\FloatBarrier

\section{Related Work}

\noindent\textbf{Long-context data construction.}
Natural long-form corpora are widely used for long-context training, but long documents are unevenly distributed across data sources~\citep{fu2024data}. Existing work constructs long contexts through query-based grouping~\citep{DBLP:conf/iclr/Gao0F025}, reference-linked document packing~\citep{zhuang2025scaling,Beyond_Rephrasing}, and hard-negative insertion~\citep{gao2025nextlong}. A complementary line develops synthetic QA supervision for retrieval and multi-evidence reasoning~\citep{li2024alr,li2024making,zhu2025generalizing,wang2026loongrl,xu2026beyond,docdancer,LongPAS,SPELL}. These approaches draw primarily on documents and webpages. ACC~\citep{su2026acc} compiles evidence from search, software-engineering, and SQL agent trajectories into shuffled long contexts for question answering, demonstrating the value of agent trajectories as a training source. We further investigate capability-targeted trajectory compilation and the links between long-context capabilities and agent atomic capabilities.

\noindent\textbf{Long-context mid-training.}
Mid-training, commonly implemented as continued pretraining (CPT), is widely used to develop long-context capabilities before downstream adaptation~\citep{mo2025mid,tu2025survey}. Alongside positional adaptation for context-window extension~\citep{ding2024longrope,shang2025longrope2,lian2025breaking}, data-centric studies examine domain mixtures and sequence-length sampling~\citep{fu2024data,PRISM}. Several training recipes combine mid-training with subsequent instruction tuning, using long-context data to strengthen context utilization~\citep{zhao2024longskywork,xu2025chatqa} or short-context data to maintain instruction-following capabilities~\citep{xu2026128k}. Dependency-aware sample selection~\citep{chen2024long} and cross-repository code contexts~\citep{paul2026octolong} further highlight the importance of long-range structure in training data.

\section{Methodology}
\label{sec:method}

\begin{figure}[!t]
    \centering
    \includegraphics[width=\linewidth]{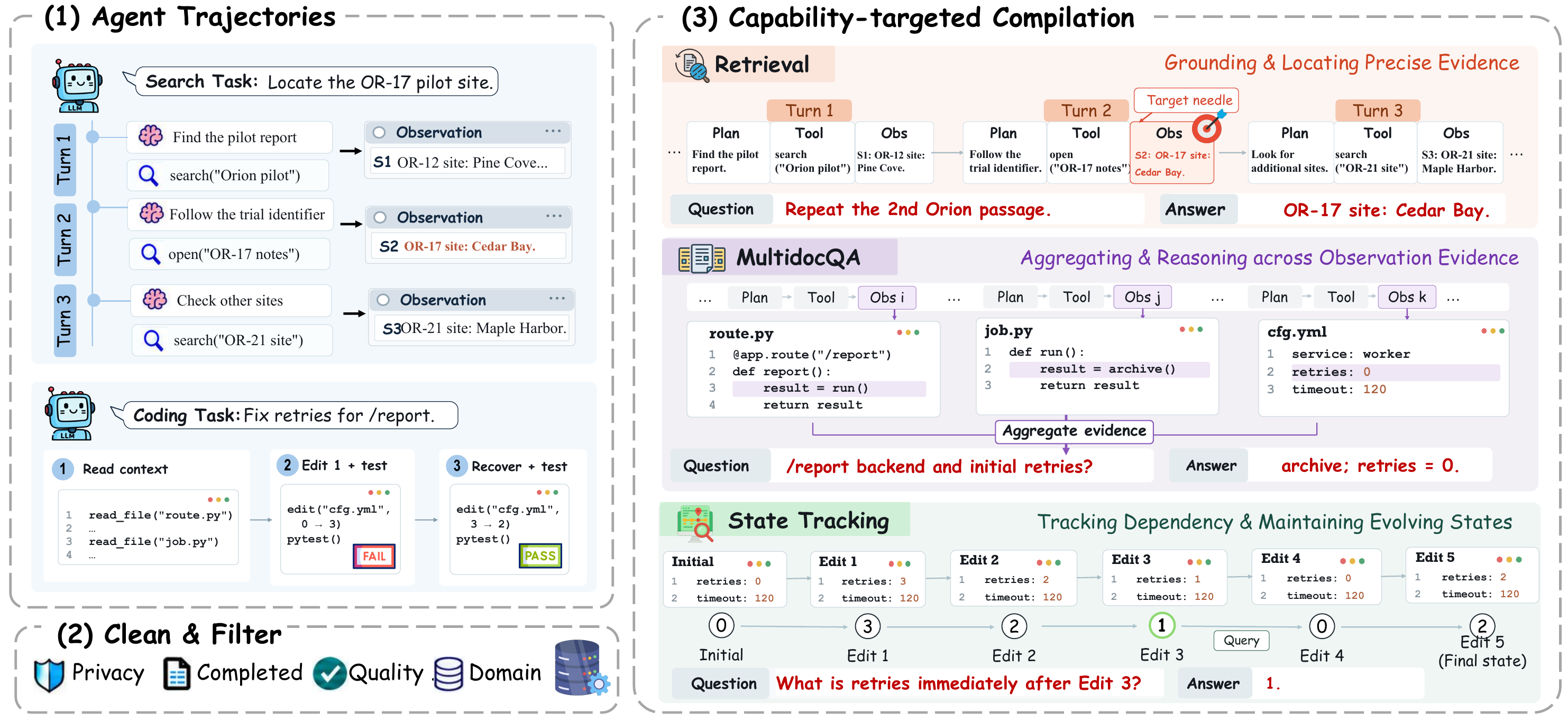}
    \caption{The \OurMODEL{} data construction pipeline. Search and coding agent trajectories are cleaned and filtered, then compiled into long-context training examples targeting evidence grounding (Retrieval), cross-evidence aggregation (MultidocQA), and temporal state maintenance (State Tracking), providing capability-targeted supervision for long-context agentic mid-training.}
    \label{fig:framework}
\end{figure}

\OurMODEL{} compiles agent trajectories into capability-targeted long-context training examples for mid-training. Figure~\ref{fig:framework} summarizes the trajectory preprocessing and compilation pipeline.

\subsection{Problem Formulation}
\label{sec:problem_formulation}
We study mid-training of a base model $p_{\theta}$ on trajectory-derived data before downstream supervised fine-tuning (SFT). A trajectory $\tau=(z_1,\ldots,z_N)$ records ordered messages, agent actions, and tool observations. Let $x=(x_1,\ldots,x_L)$ denote the token sequence of a training record and $m_t\in\{0,1\}$ indicate whether token $x_t$ is supervised. The next-token prediction objective is
\begin{equation}
    \mathcal{L}(\theta;x,m)
    = -\frac{1}{\sum_{t=1}^{L}m_t}
      \sum_{t=1}^{L}m_t\log p_{\theta}(x_t\mid x_{<t}).
    \label{eq:midtraining_objective}
\end{equation}
Unmasked mid-training on raw trajectories sets $m_t=1$ at all valid prediction positions; masked mid-training excludes system prompts, user objectives, and tool observations from the loss while retaining them as context, supervising assistant reasoning, tool calls, and responses~\citep{zhou2026deep}. \OurMODEL{} retains the next-token prediction objective with full-token supervision and reorganizes trajectory information into capability-targeted training examples under a fixed token budget.

\subsection{Trajectory Preprocessing}
\label{sec:trajectory_preprocessing}

We collect agent trajectories from real-world software-development and workplace tasks and standardize them into a unified interaction format. We replace sensitive identifiers with consistent, type-appropriate surrogates to preserve within-trajectory references and exclude harmful tasks and trajectories with unresolved privacy or confidentiality risks.

We combine rule-based checks with LLM assessment to screen for format validity, task completeness, and interaction quality. We reject traces with excessive repetition, errors, or low-value actions, while treating error recovery and progressive exploration as high-value patterns. We also remove formatting noise introduced by agent harnesses, clarify user instructions, and refine tool-response content and formatting. Retained trajectories are assigned to code or search domains according to task intent and tool-use patterns. Detailed criteria are provided in Appendix~\ref{app:trajectory_sources}.

\subsection{Capability-Targeted Compilation}
\label{sec:capability_compilation}

We target three representative long-context atomic capabilities: evidence grounding (\textit{\textbf{Retrieval}}), cross-evidence aggregation (\textit{\textbf{MultidocQA}}), and temporal state maintenance (\textit{\textbf{State Tracking}}). Detailed construction procedures are provided in Appendix~\ref{app:implementation_details}.

Given a preprocessed trajectory $\tau$, let $E(\tau)=\{o_i\}_i$ denote the evidence items extracted from its tool-call arguments and observations. A capability-specific compiler produces
\begin{equation}
    \mathcal{C}_k(\tau)=\{(c_j,q_j,y_j)\}_j,
    \qquad k\in\{\mathrm{Ret.},\mathrm{QA},\mathrm{ST}\},
    \label{eq:capability_compilers}
\end{equation}
where $c_j$ is the constructed context, $q_j$ a newly constructed question, and $y_j$ its target response derived from a teacher-generated solution. A reference answer $a_j$ is used to construct or validate the final answer in $y_j$. We omit the example index $j$ below.

\paragraph{Evidence grounding (Retrieval).}
This scheme targets precise retrieval of a specific content item from very long contexts. A single search call often returns multiple passages that share a topic but differ in their specific content. We retain the full preprocessed trajectory and treat these passages as natural \emph{needles}, selecting one as the target while the others provide topically related distractors. Questions specify both topic and within-group position, making topical matching alone insufficient to identify the requested passage. For file interactions, questions specify path, read/write operation, and occurrence. Reference answers reproduce the selected evidence item $o_{i^\star}$ as rendered, $a=\operatorname{render}(o_{i^\star})$. We check target uniqueness, answer--context alignment, and final-answer fidelity.

\paragraph{Cross-evidence aggregation (MultidocQA).}
This scheme targets multi-hop reasoning or aggregation of dispersed evidence (e.g., counting and comparison) across multiple sources. We organize file contents, retrieved documents, and execution results from different turns within a task, retaining source and version identifiers. An LLM constructs each question $q$ and reference answer $a$ from a supporting set $S\subseteq E(\tau)$. We refine questions and their supporting evidence sets to emphasize joint use of evidence. A separate solver produces a solution for each question using the complete retained evidence collection. LLM-assisted checks review question answerability, ambiguity, and output requirements, followed by checks of final-answer agreement and solution consistency.

\paragraph{Temporal state maintenance (State Tracking).}
This scheme tracks the same object through repeated updates. We align file operations with tool responses and replay successful writes and edits $e_t$ from a complete state anchor $s_0$, using $s_t=U(s_{t-1},e_t)$. Anchors come from complete writes or reads checked for completeness. For \emph{Snapshot} questions, the reference answer $a=s_{t_q}$ gives the requested state at $t_q$, covering pre-/post-edit, final, and suffix-rollback states. The context stops before the target file's first subsequent successful read, when present, to prevent copying the requested state from later observations. \emph{Provenance} questions identify the operation type and file-specific index of a queried fragment's last write, first appearance, or removal. Partial updates must be composed while preserving untouched content. We check replay success and agreement with later complete reads when available, and validate teacher-generated final answers against the reconstructed references. We exclude QA groups where prior Snapshot states directly determine later answers.

\subsection{Long-Context Mid-training}
\label{sec:long_context_midtraining}

For mid-training, we combine a fixed 30B-token background corpus with 10B tokens of compiled trajectory data, yielding approximately 40B training tokens in total. The background corpus covers code and repositories, books and papers, web data, math and STEM, and instructions and logs. Compiled variants use Retrieval, MultidocQA, or State Tracking individually, or mix them at a 3:4:3 ratio by input-token count (\OurMODEL{}-All), with matched token budgets and optimization settings.

We group compiled examples sharing a context into training records, placing the context once before $M$ QA pairs: $x=\operatorname{serialize}(c,q_1,y_1,\ldots,q_M,y_M)$. We optimize Eq.~(\ref{eq:midtraining_objective}) with $m_t=1$ at all valid prediction positions, supervising the context and all questions and target responses. Mid-training uses 256K contexts and is followed by the same downstream SFT recipe across variants. Training details and corpus statistics are provided in Appendices~\ref{app:training_details} and~\ref{app:midtraining_data_statistics}.

\section{Experimental Setup}
\label{sec:experimental_setup}

\suppressfloats[t]
\begin{table*}[t]
\centering
\definecolor{TrajAll}{HTML}{4F7A61} %
\setlength\tabcolsep{5pt}
\renewcommand{\arraystretch}{1.65}
\fontsize{8.5pt}{9.5pt}\selectfont

\resizebox{\linewidth}{!}{
\begin{tabular}{l|cc|cc|cc|cc|cc}
\toprule
\multicolumn{1}{c|}{\multirow{2}{*}{\textbf{Models}}}
& \multicolumn{2}{c|}{\textbf{LongBench V2}}
& \multicolumn{2}{c|}{\textbf{MRCR}}
& \multicolumn{2}{c|}{\textbf{GraphWalks}}
& \multicolumn{2}{c|}{\textbf{GSM-Inf}}
& \multicolumn{2}{c}{\textbf{CorpusQA}} \\
\cmidrule(lr){2-3}
\cmidrule(lr){4-5}
\cmidrule(lr){6-7}
\cmidrule(lr){8-9}
\cmidrule(lr){10-11}
& Avg@3 & BoN@3
& Avg@3 & BoN@3
& Avg@3 & BoN@3
& Avg@3 & BoN@3
& Avg@3 & BoN@3 \\
\midrule

\multicolumn{11}{c}{
  \textbf{\textit{Frontier Models}}
} \\
\midrule

Gemini-2.5-Flash-Thinking
& 58.18 & 67.00
& 59.45 & 73.52
& 41.10 & 58.41
& 25.00 & 42.67
& 77.41 & 83.28 \\

QwenLong L1.5-30B-A3B
& 45.04 & 54.69
& 55.12 & 61.51
& 39.24 & 44.16
& 39.56 & 64.33
& 80.14 & 88.45 \\

GPT-OSS-120B
& 26.15 & 38.12
& 16.66 & 26.54
& 30.79 & 35.48
& 24.00 & 42.00
& 62.31 & 75.68 \\

\midrule

\multicolumn{11}{c}{
  \textbf{\textit{Our Models}}
} \\
\midrule

Qwen3-14B
& 26.21\zerodiff & 31.94\zerodiff
& 19.16\zerodiff & 25.49\zerodiff
& 27.02\zerodiff & 32.02\zerodiff
& 3.89\zerodiff & 9.00\zerodiff
& 56.43\zerodiff & 71.73\zerodiff \\

\hspace{0.4em}- SFT \textit{(w/o Mid-training)}
& 29.34\posdiff{3.1} & 44.71\posdiff{12.8}
& 16.61\negdiff{2.6} & 26.24\posdiff{0.8}
& 24.12\negdiff{2.9} & 30.18\negdiff{1.8}
& 7.44\posdiff{3.6} & 17.67\posdiff{8.7}
& 55.93\negdiff{0.5} & 73.86\posdiff{2.1} \\

\rowcolor{TrajRetrieval!12}
\hspace{0.4em}- \OurMODEL\ \textit{(Retrieval)}
& 32.07\posdiff{5.9} & 45.31\posdiff{13.4}
& \textbf{21.60}\posdiff{2.4}
& \textbf{33.29}\posdiff{7.8}
& 29.84\posdiff{2.8} & 37.02\posdiff{5.0}
& 6.89\posdiff{3.0} & 16.67\posdiff{7.7}
& 62.92\posdiff{6.5} & 78.72\posdiff{7.0} \\

\rowcolor{TrajMultidocQA!12}
\hspace{0.4em}- \OurMODEL\ \textit{(MultidocQA)}
& \textbf{32.60}\posdiff{6.4}
& \textbf{47.50}\posdiff{15.6}
& 19.93\posdiff{0.8} & 31.60\posdiff{6.1}
& \textbf{33.46}\posdiff{6.4}
& \textbf{39.61}\posdiff{7.6}
& 7.22\posdiff{3.3}
& 18.33\posdiff{9.3}
& 63.42\posdiff{7.0}
& 78.12\posdiff{6.4} \\

\rowcolor{TrajStateTracking!12}
\hspace{0.4em}- \OurMODEL\ \textit{(State Tracking)}
& 31.87\posdiff{5.7} & 46.51\posdiff{14.6}
& 19.36\posdiff{0.2} & 30.93\posdiff{5.4}
& 32.79\posdiff{5.8} & 37.65\posdiff{5.6}
& \textbf{12.22}\posdiff{8.3}
& \textbf{26.00}\posdiff{17.0}
& \textbf{64.24}\posdiff{7.8}
& 78.42\posdiff{6.7} \\

\rowcolor{TrajAll!18}
\textbf{\OurMODEL-14B-All}
& 31.87\posdiff{5.7} & 46.91\posdiff{15.0}
& 20.18\posdiff{1.0} & 31.27\posdiff{5.8}
& 32.30\posdiff{5.3} & 38.20\posdiff{6.2}
& 6.67\posdiff{2.8} & 16.33\posdiff{7.3}
& 63.83\posdiff{7.4}
& \textbf{79.94}\posdiff{8.2} \\

\midrule

Qwen3-30B-A3B
& 23.15\zerodiff & 29.34\zerodiff
& 18.06\zerodiff & 23.86\zerodiff
& 27.29\zerodiff & 30.71\zerodiff
& 4.67\zerodiff & 10.33\zerodiff
& 56.53\zerodiff & 72.64\zerodiff \\

\hspace{0.4em}- SFT \textit{(w/o Mid-training)}
& 27.94\posdiff{4.8} & 42.32\posdiff{13.0}
& 13.55\negdiff{4.5} & 23.16\negdiff{0.7}
& 25.19\negdiff{2.1} & 32.52\posdiff{1.8}
& 5.00\posdiff{0.3} & 12.00\posdiff{1.7}
& 53.60\negdiff{2.9} & 73.86\posdiff{1.2} \\

\rowcolor{TrajAll!18}
\textbf{\OurMODEL-30B-A3B-All}
& \textbf{32.01}\posdiff{8.9} & \textbf{45.33}\posdiff{16.0}
& \textbf{19.36}\posdiff{1.3} & \textbf{28.66}\posdiff{4.8}
& \textbf{29.90}\posdiff{2.6} & \textbf{36.56}\posdiff{5.9}
& \textbf{12.44}\posdiff{7.8} & \textbf{29.33}\posdiff{19.0}
& \textbf{60.79}\posdiff{4.3} & \textbf{78.72}\posdiff{6.1} \\

\bottomrule
\end{tabular}}
\caption{Performance on five long-context benchmarks (Avg@3 and BoN@3). MRCR averages scores equally across 2/4/8-needle settings, and GraphWalks across 0--128K and 128--256K context-length intervals. Subscripts show changes relative to the corresponding Qwen3 model within each group (pp). Bold marks the best score within each group.}
\label{tab:main_results}
\vspace{-0.25cm}
\end{table*}

\paragraph{Models and Mid-training.}
We conduct controlled experiments with Qwen3-14B-Base and additionally train \OurMODEL{}-\textit{All} on Qwen3-30B-A3B-Base. Following Section~\ref{sec:long_context_midtraining}, each mixture contains 30B background tokens and 10B trajectory-derived tokens. We train with ProLong~\citep{ProLong} at 256K sequence length, using a RoPE base of $1.6{\times}10^{7}$ without additional scaling. Mid-training uses AdamW, a peak learning rate of $4{\times}10^{-5}$, cosine decay, 10\% warmup, and a global batch size of 64 packed sequences. Further training details are provided in Appendix~\ref{app:training_details}.

\paragraph{Downstream SFT.}
We construct our SFT data mixture from Nemotron-SFT-Agentic-v2, Nemotron-SFT-SWE-v3.5, and UltraChat-200K, with supervision restricted to assistant reasoning, responses, and tool calls. SFT uses 128K sequences while retaining the 256K positional configuration used for mid-training, with AdamW, a peak learning rate of $2{\times}10^{-5}$, cosine decay, 3\% warmup, and a global batch size of 32 packed sequences. Main 14B comparisons use two effective epochs; the direct-SFT baseline starts from Qwen3-14B-Base with the same positional configuration and SFT recipe. Data mixture and training details appear in Appendices~\ref{app:sft_data} and~\ref{app:training_details}.

\paragraph{Evaluation.}
To probe capabilities learned during mid-training before SFT, we adapt long-context benchmarks into ICL tasks (\textbf{Long ICL}) targeting long-context atomic capabilities. Drawing on APTBench~\citep{qin2025aptbench}, we also implement \textbf{Agent ICL} to assess base models' agent atomic capabilities. These evaluations, together with general ICL, use greedy completion or candidate likelihood scoring, with a default generation limit of 1,024 tokens. After SFT, we evaluate long-context, search, tool-use, and coding agent benchmarks. Our downstream evaluation runs use task-level defaults of temperature 0.7, top-$p$ 1.0, and a 65,536-token output cap per model call, subject to benchmark-specific settings. We report average and best-of-three metrics (Avg@3 and BoN@3), with score transformations and repeated-trial aggregation defined in Appendix~\ref{app:evaluation_protocols}. More details about baselines and benchmarks are presented in Appendices~\ref{app:benchmarks_baselines} and~\ref{app:general_icl}.

\section{Experimental Results}

\subsection{Main Results}
Following mid-training and SFT, we evaluate \OurMODEL{} on Qwen3-14B and Qwen3-30B-A3B against SFT-only counterparts and frontier LLMs. We assess long-context performance and downstream search, tool-use, and coding tasks, comparing the three specialized 14B variants with their mixture to examine task-specific strengths and broader gains. Additional results are provided in Appendix~\ref{app:additional_experiments}.

\paragraph{Capability-Targeted Compilation Improves Long-Context Performance.}
Table~\ref{tab:main_results} shows distinct long-context strengths across the three compilation schemes. Among the Qwen3-14B variants, \textit{Retrieval} leads on MRCR, improving Avg@3 by 4.99 percentage points over SFT without mid-training. \textit{MultidocQA} performs best on LongBench V2 and GraphWalks, while \textit{State Tracking} achieves the strongest GSM-Infinite results, with a 4.78-point Avg@3 gain over the same baseline. These task-dependent differences are consistent with the distinct information dependencies emphasized during compilation. Combining the three capabilities in \OurMODEL-\textit{All} improves both Avg@3 and BoN@3 over SFT-only on nearly all benchmarks for Qwen3-14B and Qwen3-30B-A3B. These findings suggest that compiling agent trajectories into tasks targeting atomic capabilities can improve general long-context performance, with each scheme exhibiting distinct task-specific strengths.

\paragraph{Long-Context Gains Transfer to Downstream Agentic Tasks.}
Table~\ref{tab:agent_results_avg} shows improvements across search, tool use, and coding under the same downstream SFT recipe. For Qwen3-14B, each specialized compilation scheme outperforms SFT without mid-training on all 12 agent benchmarks. Combining the three schemes in \OurMODEL-14B-All yields the strongest SWE-Verified and SWE-Pro results among the 14B and 30B variants. These results suggest that capability-targeted long-context mid-training supports downstream transfer across search, tool use, and coding. The benefits extend beyond a fixed correspondence between compilation schemes and agent categories, suggesting that shared long-context capabilities provide a transferable foundation for diverse agentic tasks.

\begin{table*}[!t]
\centering
\definecolor{TrajAll}{HTML}{4F7A61} %
\setlength\tabcolsep{2.0pt}
\renewcommand{\arraystretch}{1.65}
\fontsize{8.5pt}{9.5pt}\selectfont

\resizebox{\linewidth}{!}{
\begin{tabular}{l|ccccc|ccc|cccc}
\toprule
\multicolumn{1}{c|}{\multirow{2}{*}{\textbf{Models}}}
& \multicolumn{5}{c|}{\textbf{Search}}
& \multicolumn{3}{c|}{\textbf{Tool Use}}
& \multicolumn{4}{c}{\textbf{Coding}} \\
\cmidrule(lr){2-6}
\cmidrule(lr){7-9}
\cmidrule(lr){10-13}

& \textbf{BC}
& \textbf{BC-ZH}
& \textbf{WideSearch}
& \textbf{FSC-V2}
& \textbf{SealQA}
& \textbf{MCP-Atlas}
& \textbf{SkillsBench}
& \textbf{WebArena}
& \textbf{TB 2.1}
& \textbf{NL2Repo}
& \textbf{SWE-Verified}
& \textbf{SWE-Pro} \\

\midrule

\multicolumn{13}{c}{\textbf{\textit{Frontier Models}}} \\
\midrule

Gemini-2.5-Flash-Thinking
& 12.72 & 27.89 & 30.84 & 34.48 & 24.02 & 19.87 & 7.41 & 39.43 & 17.75 & 12.40 & 48.60 & 24.08 \\

QwenLong L1.5-30B-A3B
& 2.98 & 11.07 & 35.27 & 20.36 & 16.58
& 32.73 & 1.39 & 27.17
& 1.87 & 7.71 & 31.60 & 10.94 \\

GPT-OSS-120B
& 25.62 & 24.22 & 36.46 & 41.09 & 22.52 & 39.40 & - & 31.40 & 12.36 & 6.88 & 53.00 & 27.13 \\

\midrule

\multicolumn{13}{c}{\textbf{\textit{Our Models}}} \\
\midrule

Qwen3-14B
& 1.87 & 10.45 & 29.91 & 22.81 & 13.87
& 11.93 & 0.28 & 19.80
& 0.37 & 5.59 & 18.07 & 4.29 \\

\hspace{0.4em}- SFT \textit{(w/o Mid-training)}
& 12.53 & 18.27 & 24.96 & 35.09 & 24.32
& 42.40 & 6.48 & 27.32
& 7.12 & 6.23 & 42.13 & 24.21 \\

\rowcolor{TrajRetrieval!12}
\hspace{0.4em}- \OurMODEL\ \textit{(Retrieval)}
& 13.14 & 18.75 & 30.39 & 40.66 & 26.13
& 45.80 & 9.26 & \textbf{30.32}
& \textbf{15.73} & \textbf{17.66} & 50.87 & 32.15 \\

\rowcolor{TrajMultidocQA!12}
\hspace{0.4em}- \OurMODEL\ \textit{(MultidocQA)}
& \textbf{13.93} & 18.34 & 30.48 & 42.66 & 24.68
& \textbf{46.73} & 6.94 & 29.88
& 9.36 & 15.71 & 52.13 & 35.11 \\

\rowcolor{TrajStateTracking!12}
\hspace{0.4em}- \OurMODEL\ \textit{(State Tracking)}
& 12.72 & \textbf{19.58} & \textbf{32.98} & 40.97 & \textbf{31.35}
& 45.00 & \textbf{12.04} & 28.42
& 14.61 & 15.40 & 53.07 & 35.89 \\

\rowcolor{TrajAll!18}
\textbf{\OurMODEL-14B-All}
& 11.77 & 19.38 & 30.42 & \textbf{43.53} & 24.32
& 44.00 & 11.11 & 29.48
& 14.23 & 15.03 & \textbf{54.40} & \textbf{36.43} \\

\midrule

Qwen3-30B-A3B
& 2.00 & 10.59 & 26.62 & 20.46 & 14.77
& 20.53 & 0.93 & 12.32
& 1.50 & 7.08 & 25.47 & 9.48 \\

\hspace{0.4em}- SFT \textit{(w/o Mid-training)}
& 10.27 & 16.54 & 22.51 & 38.31 & \textbf{25.41} & 37.67 & 4.63 & 25.49 & 1.87 & 9.57 & 36.53 & 23.07 \\

\rowcolor{TrajAll!18}
\textbf{\OurMODEL-30B-A3B-All}
& \textbf{11.35} & \textbf{20.07} & \textbf{28.16} & \textbf{41.69} & 25.23
& \textbf{40.53} & \textbf{7.41} & \textbf{26.13}
& \textbf{12.36} & \textbf{13.79} & \textbf{51.07} & \textbf{34.34} \\

\bottomrule
\end{tabular}
}

\caption{
Performance on search, tool-use, and coding agent benchmarks (Avg@3). Bold marks the best score within each group. Benchmark details and BoN@3 results are in Appendix~\ref{app:agent_evaluation} and~\ref{app:main_results_agent}.
}
\label{tab:agent_results_avg}
\vspace{-0.25cm}
\end{table*}

\subsection{Ablation Study}

\suppressfloats[t]

Table~\ref{tab:ablation_midtraining} compares data construction under matched mid-training token budgets, a fixed background corpus, and the same downstream SFT recipe. Search trajectories outperform code trajectories on Oolong-Synth and SkillsBench, whereas code trajectories are stronger on GraphWalks and the two SWE benchmarks. Capability-targeted compilation yields broader improvements. \OurMODEL-\textit{All} outperforms both unmasked raw trajectories (\textit{Combined Trajectories}) and masked raw trajectories (\textit{Combined Trajectories (with Mask)}) on all seven benchmarks. Relative to the unmasked baseline, it improves Avg@3 by 2.03 points on CorpusQA and 2.66 points on FSC-V2, with additional gains on tool-use and coding tasks. These results suggest that organizing trajectory information into capability-targeted long-context tasks provides benefits beyond source mixing or loss masking alone.

\begin{table*}[!h]
\centering
\definecolor{TrajAll}{HTML}{4F7A61} %
\setlength\tabcolsep{8pt}
\renewcommand{\arraystretch}{1.65}
\fontsize{8.5pt}{9.5pt}\selectfont

\resizebox{\linewidth}{!}{
\begin{tabular}{l|ccc|cccc}
\toprule
\multicolumn{1}{c|}{\multirow{2}{*}{\textbf{Mid-training Variant}}}
& \multicolumn{3}{c|}{\textbf{Long-Context}}
& \multicolumn{4}{c}{\textbf{Agent}} \\
\cmidrule(lr){2-4}
\cmidrule(lr){5-8}
& \textbf{Oolong-Synth}
& \textbf{GraphWalks}
& \textbf{CorpusQA}
& \textbf{FSC-V2}
& \textbf{SkillsBench}
& \textbf{SWE-Verified}
& \textbf{SWE-Pro} \\
\midrule

\rowcolor{TrajAll!18}
\OurMODEL-\textit{All}
& 58.50\zerodiff
& 32.30\zerodiff
& 63.83\zerodiff
& 43.53\zerodiff
& 11.11\zerodiff
& 54.40\zerodiff
& 36.43\zerodiff \\
\midrule

\textit{SFT only}
& 53.17\negdiff{5.33}
& 24.12\negdiff{8.18}
& 55.93\negdiff{7.90}
& 35.09\negdiff{8.44}
& 6.48\negdiff{4.63}
& 42.13\negdiff{12.27}
& 24.21\negdiff{12.22} \\

\textit{Search Trajectories}
& 57.33\negdiff{1.17}
& 31.58\negdiff{0.72}
& 61.80\negdiff{2.03}
& 40.05\negdiff{3.48}
& 9.72\negdiff{1.39}
& 50.27\negdiff{4.13}
& 31.37\negdiff{5.06} \\

\textit{Code Trajectories}
& 55.33\negdiff{3.17}
& 32.77\posdiff{0.47}
& 61.80\negdiff{2.03}
& 39.49\negdiff{4.04}
& 8.80\negdiff{2.31}
& 52.40\negdiff{2.00}
& 34.38\negdiff{2.05} \\

\textit{Combined Trajectories}
& 56.67\negdiff{1.83}
& 31.38\negdiff{0.92}
& 61.80\negdiff{2.03}
& 40.87\negdiff{2.66}
& 9.26\negdiff{1.85}
& 53.60\negdiff{0.80}
& 35.93\negdiff{0.50} \\

\textit{Combined Trajectories (with Mask)}
& 55.00\negdiff{3.50}
& 31.25\negdiff{1.05}
& 60.99\negdiff{2.84}
& 42.20\negdiff{1.33}
& 6.48\negdiff{4.63}
& 53.20\negdiff{1.20}
& 33.88\negdiff{2.55} \\

\bottomrule
\end{tabular}
}

\caption{Ablation Study (Avg@3) of the mid-training data construction on long-context and agent benchmarks. All variants use the same Qwen3-14B backbone, downstream SFT recipe, and evaluation protocol. The trajectory baselines use raw search trajectories, raw code trajectories, or their combination, with an additional masked variant. All mid-training runs use the same token budget.}
\label{tab:ablation_midtraining}
\vspace{-0.25cm}
\end{table*}

\FloatBarrier
\subsection{Analysis}
\label{sec:experimental_analysis}

\subsubsection{Capability-Aligned ICL Learning Dynamics}

\suppressfloats[t]
\begin{figure*}[t]
    \centering
    \includegraphics[width=1\textwidth]{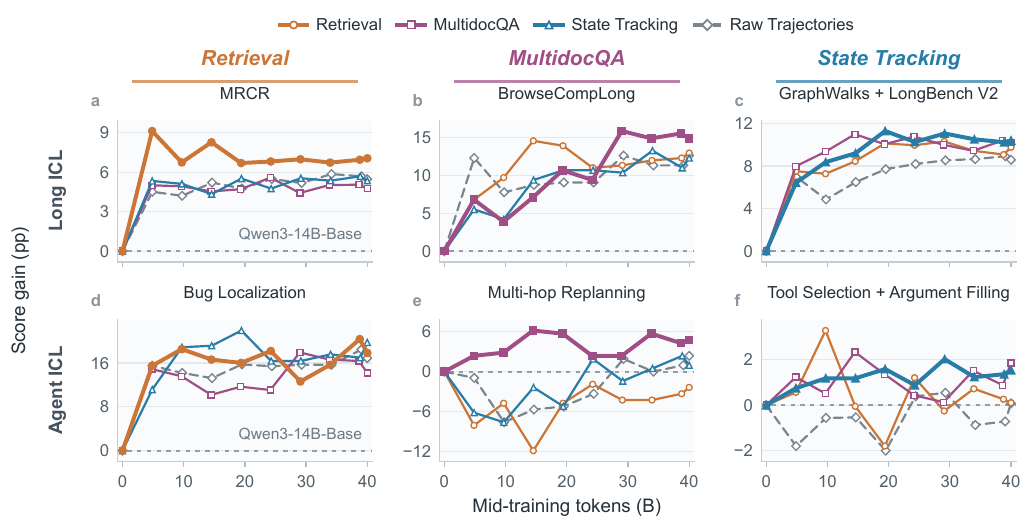}
    \caption{ICL score gains over Qwen3-14B-Base on long-context (top) and agent (bottom) tasks during mid-training (pp). Training steps are scaled to 40B tokens per run.}
    \label{fig:capability_aligned_icl_dynamics}
\end{figure*}

Figure~\ref{fig:capability_aligned_icl_dynamics} shows that long-context ICL gains generally rise before leveling off. \textit{Retrieval} reaches its highest observed MRCR gain at the first evaluated checkpoint (approximately 5B tokens), whereas \textit{MultidocQA} on BrowseCompLong and \textit{State Tracking} on GraphWalks + LongBench V2 peak later. On Agent ICL, \textit{MultidocQA} leads multi-hop replanning throughout training, while \textit{State Tracking} consistently outperforms Raw Trajectories on the tool-selection/argument-filling composite.

These results suggest that \textbf{targeted trajectory supervision yields overlapping gains across long-context and agent tasks, but with task-dependent learning dynamics}. \textit{Retrieval} improves MRCR sharply early in training, whereas \textit{MultidocQA} on BrowseCompLong and \textit{State Tracking} on GraphWalks + LongBench V2 achieve their largest observed gains later, despite intermediate fluctuations. The timing of these gains also differs between long-context tasks and agent ICL.

\subsubsection{Long--Agent Associations across Atomic Capabilities}

We examine how mid-training ICL gains on proxies for the three long-context atomic capabilities correlate with gains in individual agent atomic capabilities (Figure~\ref{fig:long_agent_icl_correlation}). This Spearman analysis uses five training variants (Retrieval, MultidocQA, State Tracking, All, and Raw Trajectories), with gains relative to unmasked Raw Trajectories. Final-checkpoint ICL gain profiles for the individual training variants are reported in Appendix~\ref{app:icl_gain_profiles}.

\begingroup
\setlength{\columnsep}{9pt}
\setlength{\intextsep}{3pt}
\begin{wrapfigure}{R}{0.56\textwidth}
    \centering
    \includegraphics[width=\linewidth,trim=0bp 4bp 8bp 5bp,clip]{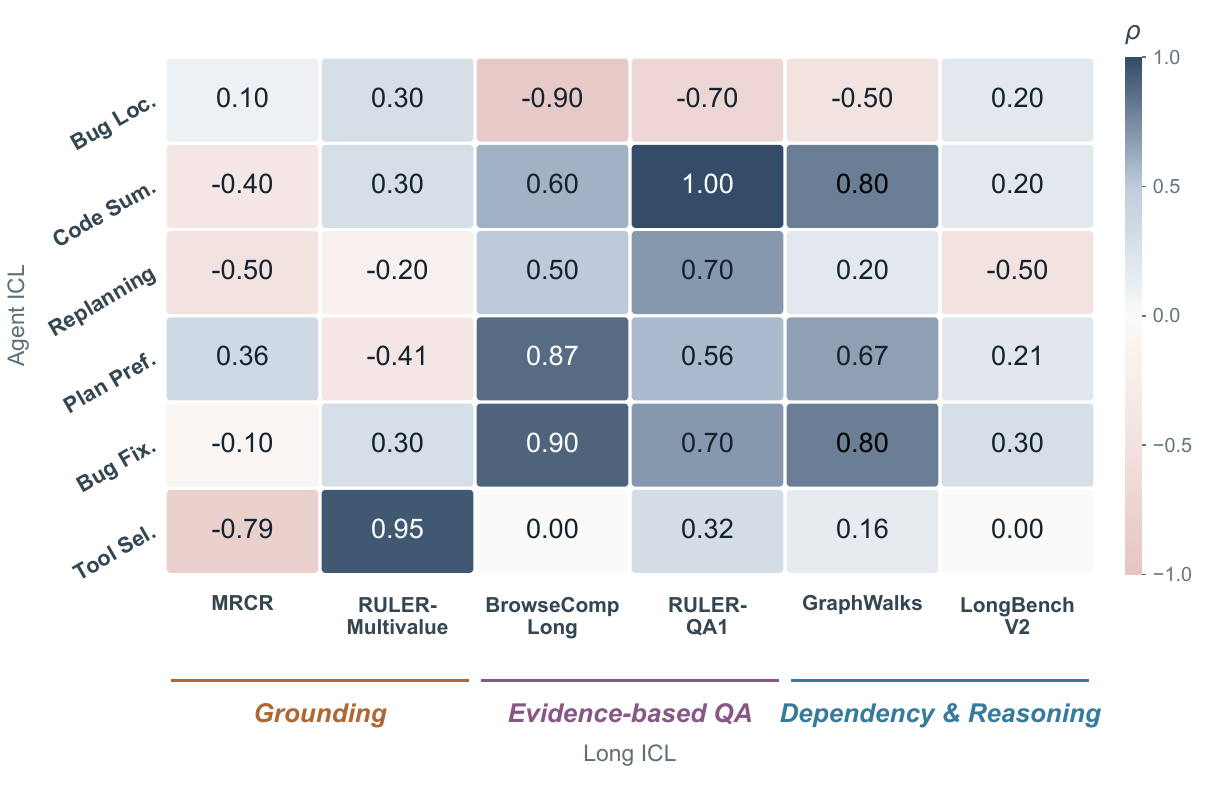}
    \captionsetup{font=small,skip=5pt}
    \caption{Spearman correlations between long-context and agent ICL gains across five Qwen3-14B variants at final mid-training checkpoints, relative to unmasked Raw Trajectories.}
    \label{fig:long_agent_icl_correlation}
    \vspace{-\baselineskip}
\end{wrapfigure}

Within the \textit{Grounding} group, RULER-Multivalue is most positively associated with Tool Selection, consistent with selecting task-relevant information from competing candidates.

The \textit{Evidence-based QA} group, comprising BrowseCompLong and RULER-QA1, shows more consistent positive associations with Code Summary, Bug Fixing, Planning Preference, and Multi-hop Replanning across both benchmarks. These patterns are consistent with shared demands for using and aggregating contextual evidence when interpreting code, evaluating repairs, or choosing actions based on interaction history.

Within the \textit{Dependency \& Reasoning} group, GraphWalks correlates strongly with Code Summary and Bug Fixing (both $\rho=0.80$), followed by Planning Preference ($\rho=0.67$). This pattern suggests shared demands for tracing dependencies and understanding program behavior. Together, \textbf{evidence-based QA and dependency reasoning exhibit overlapping associations with code understanding and planning}, while the weaker or divergent patterns on MRCR and LongBench V2 suggest that these relationships depend on the specific information-processing demands of each benchmark.

\par
\ifnum\value{WF@wrappedlines}>1\relax
    \vspace{\dimexpr\value{WF@wrappedlines}\baselineskip\relax}
\fi
\WFclear
\endgroup

\FloatBarrier
\subsubsection{Capability-Specific Gains after Downstream SFT}
\begin{figure*}[htbp]
    \centering
    \includegraphics[width=\textwidth]{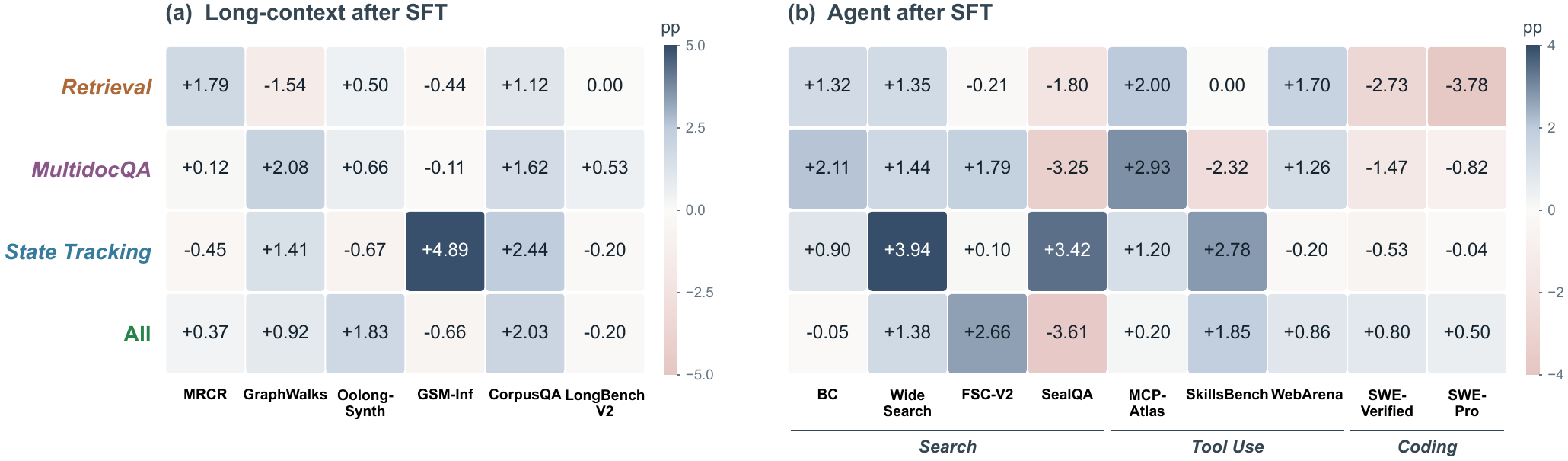}
    \caption{Post-SFT Avg@3 gains of Qwen3-14B variants on long-context and agent benchmarks, relative to unmasked Raw Trajectories under the same SFT recipe. Panels use separate color scales.}
    \label{fig:sft_capability_gains}
\end{figure*}

Figure~\ref{fig:sft_capability_gains} compares the compiled variants with unmasked Raw Trajectories under the same downstream SFT recipe. \textit{Retrieval} yields the largest gains on MRCR ($+1.79$ pp) and WebArena, alongside improvements on MCP-Atlas, but regresses on GraphWalks and both SWE benchmarks. \textbf{Its retrieval advantage therefore coexists with benefits on selected interactive tasks}, without extending uniformly to dependency reasoning or software engineering.

\textit{MultidocQA} leads on GraphWalks and LongBench V2, while also improving CorpusQA. Its strongest agent gains occur on BrowseComp and MCP-Atlas ($+2.93$ pp). These paired advantages are consistent with shared demands for \textbf{combining contextual evidence and tracing dependencies in long-context reasoning, information seeking, and multi-step tool composition}.

\textit{State Tracking} achieves the largest gain on GSM-Inf ($+4.89$ pp) and CorpusQA. On agent tasks, it leads on WideSearch, SealQA, and SkillsBench. The CorpusQA--WideSearch pairing suggests a shared need to consolidate many facts into a coherent answer, while the GSM-Inf gain is consistent with tracking intermediate quantities through dependent computations. Thus, \textbf{state-oriented supervision may support aggregation and structured reasoning beyond explicitly temporal tasks}.

\FloatBarrier

\subsubsection{\OurMODEL{} across Training Stages}
\label{sec:training_stage_analysis}

We further analyze whether \OurMODEL{}-compiled data also benefit direct SFT without mid-training. We convert them to dialogue format and mix them with the original SFT data at 3:7, preserving the total SFT token amount. Table~\ref{tab:training_stage_comparison} shows gains over \textit{None} on all six benchmarks, indicating that \textbf{capability-targeted compilation of \OurMODEL{} remains useful under assistant-only SFT stage}. Under the same two-epoch direct-SFT recipe, 3:7 mixture ratio, and token budget, \textbf{\OurMODEL{} outperforms ACC~\citep{su2026acc} on most long-context benchmarks and all three evaluated agent benchmarks}, including a $3.89$ pp advantage on SWE-Pro.

Across one to three SFT epochs after mid-training, long-context gains over \textit{None} remain broadly stable, while agent performance shows modest overall improvement with additional epochs. These results suggest that extending SFT preserves the long-context benefits of our mid-training recipe while providing incremental improvements in downstream agent performance.

\begin{table*}[htbp]
\centering
\setlength\tabcolsep{4pt}
\renewcommand{\arraystretch}{1.45}
\fontsize{9pt}{10.5pt}\selectfont

\resizebox{0.9\linewidth}{!}{%
\begin{tabular}{c|c|cccccc}
\toprule
\multirow{2}{*}{\textbf{Setting}}
& \multirow{2}{*}{\textbf{SFT Epoch}}
& \multicolumn{3}{c}{\textbf{Long-Context}}
& \multicolumn{3}{c}{\textbf{Agent}} \\
\cmidrule(lr){3-5}
\cmidrule(lr){6-8}
&
& \textbf{Oolong-Synth}
& \textbf{GraphWalks}
& \textbf{CorpusQA}
& \textbf{BC}
& \textbf{NL2Repo}
& \textbf{SWE-Pro} \\
\midrule

None
& 2
& 53.17\zerodiff
& 24.12\zerodiff
& 55.93\zerodiff
& 12.53\zerodiff
& 6.23\zerodiff
& 24.21\zerodiff \\

ACC (SFT)
& 2
& 56.00\posdiff{2.83}
& 30.62\posdiff{6.50}
& 57.89\posdiff{1.96}
& 10.16\negdiff{2.37}
& 7.00\posdiff{0.77}
& 27.22\posdiff{3.01} \\

\rowcolor{gray!8}
\OurMODEL{} (SFT)
& 2
& 56.83\posdiff{3.66}
& 28.04\posdiff{3.92}
& 59.27\posdiff{3.34}
& 13.56\posdiff{1.03}
& 9.40\posdiff{3.17}
& 31.11\posdiff{6.90} \\
\midrule

& 1
& 55.83\posdiff{2.66}
& 31.64\posdiff{7.52}
& \textbf{64.44}\posdiff{8.51}
& 11.48\negdiff{1.05}
& 14.16\posdiff{7.93}
& 34.15\posdiff{9.94} \\

\rowcolor{TrajAll!16}
& 2
& \textbf{58.50}\posdiff{5.33}
& \textbf{32.30}\posdiff{8.18}
& 63.83\posdiff{7.90}
& 11.77\negdiff{0.76}
& \textbf{15.03}\posdiff{8.80}
& 36.43\posdiff{12.22} \\

\multirow{-3}{*}{\shortstack{\OurMODEL{}\\(Mid-train+SFT)}}
& 3
& \textbf{58.50}\posdiff{5.33}
& 32.02\posdiff{7.90}
& 63.22\posdiff{7.29}
& \textbf{14.67}\posdiff{2.14}
& 14.81\posdiff{8.58}
& \textbf{37.21}\posdiff{13.00} \\
\bottomrule
\end{tabular}
}

\caption{Training-stage and SFT-epoch comparisons on Qwen3-14B (Avg@3). \textit{None} denotes SFT on the original data only. Subscripts show changes from \textit{None} (pp); bold marks column maxima.}
\label{tab:training_stage_comparison}
\end{table*}

\FloatBarrier

\Needspace{12\baselineskip}
\section{Conclusion}
\label{sec:conclusion}

We presented \OurMODEL{}, a framework that compiles agent trajectories into long-context supervision targeting evidence grounding, cross-evidence aggregation, and temporal state maintenance. Experiments on Qwen3-14B and 30B show gains across long-context and agent benchmarks, comparing with both raw and masked trajectories. Our analyses reveal task-dependent learning dynamics and correlations between long-context performance and agent atomic capabilities, with evidence-based QA and dependency reasoning associated with code understanding and planning. State-oriented supervision may also support aggregation and structured reasoning beyond temporal tasks. Together, these findings support organizing trajectory information around long-context capabilities as a basis for effective agentic mid-training rather than being left entirely to post-training.

\bibliography{iclr2027_conference}
\bibliographystyle{plainnat}

\clearpage
\appendix
\section{Dataset Details}
\label{app:data_statistics}

\subsection{Trajectory Sources and Cleaning}
\label{app:trajectory_sources}

\paragraph{Sources and Permissions.}
Our source corpus consists of agent trajectories generated while carrying out real-world user tasks in software development and workplace settings. Each trajectory records user requests, assistant responses, and tool interactions. All data used in this study were collected with users' knowledge and authorization, with the necessary permissions and licenses obtained for their use in this research.

\paragraph{Privacy Protection.}
We apply three complementary forms of privacy protection: user-level de-identification, organization-level exclusion, and task-level semantic privacy screening.

\emph{User-level de-identification.} The replacement pipeline targets three groups of sensitive information:
\begin{itemize}[label={$\bullet$},leftmargin=1.2em,itemsep=2pt,topsep=3pt,parsep=0pt]
    \item \textbf{Personal identifiers:} user identifiers, names, email addresses, phone numbers, identity numbers, and payment-card numbers.
    \item \textbf{Authentication material:} API keys, access tokens, embedded passwords, private keys, and database connection strings.
    \item \textbf{Environment-specific identifiers:} personal filesystem paths, version-control identity fields, IP addresses, and internal domains.
\end{itemize}
Detection combines pattern matching with contextual checks, allowlists, and format or checksum validation where applicable. It operates on message text, including tool observations, as well as reasoning fields and string values in structured tool-call arguments. Detected values are replaced with randomly generated, type-appropriate synthetic values. Repeated values of the same category within a trajectory receive consistent replacements, preserving references across messages and tool interactions.

\emph{Organization-level exclusion.} We discard entire trajectories containing prominent company identifiers or involving internal documents. The unit of removal is the complete trajectory, including its user requests, assistant responses, and tool observations, so that associated organizational content is excluded together with the identifying material.

\emph{Task-level semantic privacy screening.} Sensitive information can also be conveyed through task content after explicit identifiers are removed~\citep{mireshghallah2024trust}. We therefore use an LLM to assess whether a task involves sensitive personal information or confidential company projects. Trajectories flagged by this semantic screening are discarded in full. This criterion addresses privacy and confidentiality in the task's meaning and context, complementing explicit identifier replacement and organization-level filtering.

\paragraph{Task Safety.}
Separately, an LLM assesses task objectives for harmful intent, including malicious cyberattacks, phishing, fraud, and dangerous weapon-related instructions. It distinguishes such objectives from routine software development and data processing. Trajectories labeled unsafe are excluded in full. Task safety and privacy are separate exclusion criteria: a task with a benign objective may still contain sensitive personal or organizational information.

\paragraph{Cleaning and Filtering.}
After standardizing the collected trajectories into a unified interaction format, we apply structural and quality checks covering format validity, task completeness, and interaction quality (Table~\ref{tab:trajectory_cleaning}). These checks remove malformed, incomplete, and low-value trajectories while retaining informative patterns such as error recovery and progressive exploration.

\begin{table}[htbp]
    \centering
    \small
    \setlength{\tabcolsep}{5pt}
    \renewcommand{\arraystretch}{1.12}
    \setlist[itemize]{nosep,leftmargin=1em,label={$\bullet$}}
    \begin{tabular}{@{}p{0.20\linewidth}p{0.13\linewidth}p{\dimexpr0.67\linewidth-20pt\relax}@{}}
        \toprule
        \textbf{Criterion} & \textbf{Assessment} & \textbf{Selection principle} \\
        \midrule
        Format validity & Rules &
        \begin{minipage}[t]{\linewidth}
        \begin{itemize}
            \item Nonempty user request and assistant text or actions.
            \item $\geq 5$ non-system messages; $\geq 1$ matched tool-call/response pair.
            \item Remove agent-harness boilerplate and tool-output formatting noise.
        \end{itemize}
        \end{minipage} \\
        \addlinespace[5pt]
        Task completeness & Rules + LLM &
        \begin{minipage}[t]{\linewidth}
        \begin{itemize}
            \item Remove dangling terminal calls; reject LLM-labeled incomplete workflows.
            \item Allow completed attempts with explicit external constraints.
        \end{itemize}
        \end{minipage} \\
        \addlinespace[5pt]
        Quality & Rules &
        \begin{minipage}[t]{\linewidth}
        \begin{itemize}
            \item Reject $\geq 4$ consecutive calls with identical tools and arguments.
            \item Reject error-marked responses exceeding 70\% of all tool responses.
        \end{itemize}
        \end{minipage} \\
        \addlinespace[3pt]
        & LLM &
        \begin{minipage}[t]{\linewidth}
        \begin{itemize}
            \item High-value patterns: error recovery and progressive exploration.
            \item Low-value patterns: unproductive loops, dead ends, and irrelevant actions.
            \item Ratios: summed annotated segment lengths / total nodes;
            reject low-value ratio $>50\%$.
            \item Clarify user instructions and refine tool-response content.
        \end{itemize}
        \end{minipage} \\
        \bottomrule
    \end{tabular}
    \caption{Trajectory cleaning criteria.}
    \label{tab:trajectory_cleaning}
\end{table}

\paragraph{Domain Classification.}
We organize the trajectories used in this study into two task domains: \emph{code} and \emph{search}. Assignment combines tool-use patterns with task intent; auxiliary calls such as planning and memory management are excluded from the tool counts.
\begin{itemize}[label={$\bullet$},leftmargin=1.2em,itemsep=3pt,topsep=3pt,parsep=0pt]
    \item \textbf{Code.} We identify coding trajectories from the prevalence of code reading, editing, and execution tools, supplemented by agent-scaffold context. A keyword check distinguishes software work from document production. Tasks include feature development, refactoring, bug fixing and debugging, testing, codebase analysis, maintenance automation, configuration management, and DevOps.
    \item \textbf{Search.} We select trajectories with at least two web-retrieval calls accounting for at least 40\% of non-auxiliary calls. Retained tasks include information retrieval and question answering, as well as research and report synthesis from retrieved evidence.
\end{itemize}

\FloatBarrier
\Needspace{10\baselineskip}
\subsection{Mid-training Data Statistics}
\label{app:midtraining_data_statistics}

Each mid-training variant combines an approximately 30B-token background corpus with a 10B-token component of original trajectories or \OurMODEL{} data. We report lengths of complete training records before packing, using the Qwen3-14B-Base tokenizer without adding BOS or EOS tokens. Repeated sampling occurrences are counted separately; k and B denote $10^3$ and $10^9$ tokens.

\paragraph{Raw Trajectory Statistics.}
\label{app:raw_trajectory_statistics}
Table~\ref{tab:cpt_trajectory_statistics} summarizes the code and search trajectory corpora, whose mean lengths are 82.59k and 65.91k tokens, respectively. User turns count user-role messages, and tool calls count individual calls issued by the assistant. Before packing, the data-level loss mask retains supervision on only 22.51\% and 9.82\% of tokens, respectively; the remaining tokens provide context without direct supervision in the masked variant.

\paragraph{\OurMODEL{} Data Statistics.}
\label{app:cpt_length_distributions}
Figure~\ref{fig:cpt_length_distribution} compares the length distributions of original trajectories and the three \OurMODEL{} data components using 2k-token bins over 2k--300k, with each curve normalized by its corpus's full record count.

\par\medskip
\noindent\begin{minipage}{\linewidth}
    \begin{minipage}[t]{0.46\linewidth}
        \vspace{0pt}
        \captionsetup{font=footnotesize,justification=raggedright,singlelinecheck=false,hypcap=false,skip=6pt}
        \centering
\footnotesize
\setlength{\tabcolsep}{3pt}
\renewcommand{\arraystretch}{1.16}
\begin{tabular*}{\linewidth}{@{}l@{\extracolsep{\fill}}rr@{}}
\toprule
\textbf{Statistic} & \textbf{Code} & \textbf{Search} \\
\midrule
Training records & 127,153 & 159,017 \\
Total tokens (B) & 10.502 & 10.481 \\
Mean length (k) & 82.59 & 65.91 \\
\addlinespace[3pt]
Mean user turns & 4.82 & 1.94 \\
Mean tool calls & 51.9 & 16.4 \\
\midrule
Supervised tokens (B) & 2.364 & 1.029 \\
Supervision density (\%) & \textbf{22.51} & \textbf{9.82} \\
\bottomrule
\end{tabular*}

        \par
        \captionof{table}{Raw code and search trajectory statistics.}
        \label{tab:cpt_trajectory_statistics}
    \end{minipage}\hfill
    \begin{minipage}[t]{0.52\linewidth}
        \vspace{0pt}
        \centering
        \includegraphics[width=\linewidth]{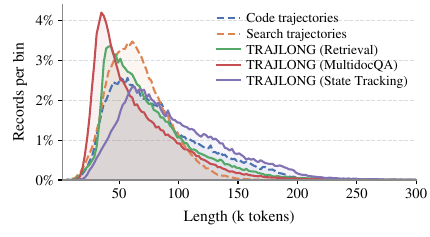}
        \par\vspace{3pt}
        \captionsetup{font=footnotesize,justification=raggedright,singlelinecheck=false,hypcap=false,skip=0pt}
        \captionof{figure}{Trajectory and \OurMODEL{} length distributions.}
        \label{fig:cpt_length_distribution}
    \end{minipage}
\end{minipage}
\par\medskip

\paragraph{Shared Mid-training Corpus.}
The approximately 30B-token background corpus contains about 5.6 million training records, with a mean length of 5.50k tokens. About 76.2\% of records are shorter than 2k tokens. Table~\ref{tab:background_mixture} reports the mixture composition and token budgets, and Figure~\ref{fig:midtraining_tsne} visualizes the compiled data alongside the background corpus.

\par\medskip
\noindent\begin{minipage}{\linewidth}
\centering
\begingroup
\begin{minipage}[t]{0.46\linewidth}
\vspace{0pt}
\centering
\captionsetup{font=footnotesize,justification=raggedright,singlelinecheck=false,hypcap=false,skip=6pt}
\footnotesize
\setlength{\tabcolsep}{3pt}%
\renewcommand{\arraystretch}{1.35}%
\begin{tabular*}{0.88\linewidth}{@{}l@{\extracolsep{\fill}}rr@{}}
\toprule
\textbf{Category} & \textbf{Tokens (B)} & \textbf{Ratio} \\
\midrule
Code \& Repos & 9.4 & 23.50\% \\
Books \& Papers & 10.1 & 25.25\% \\
Web & 5.1 & 12.75\% \\
Math \& STEM & 1.8 & 4.50\% \\
Instructions \& Logs & 3.6 & 9.00\% \\
\midrule
\shortstack[l]{Compiled Agent\\Trajectories} & 10.0 & 25.00\% \\
\midrule
\textbf{Total} & \textbf{40.0} & \textbf{100.00\%} \\
\bottomrule
\end{tabular*}
\par
\captionof{table}{Data composition and token budgets of the \OurMODEL{} mid-training mixture.}
\label{tab:background_mixture}
\end{minipage}\hfill
\begin{minipage}[t]{0.52\linewidth}
\vspace{0pt}
\centering
\includegraphics[width=\linewidth,height=2.12in,keepaspectratio]{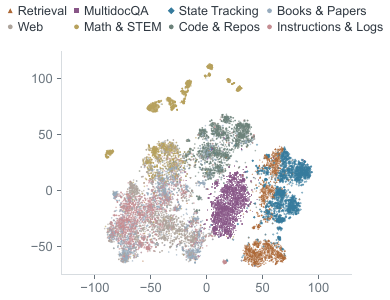}
\par\vspace{3pt}
\captionsetup{font=footnotesize,justification=raggedright,singlelinecheck=false,hypcap=false,skip=0pt}
\captionof{figure}{t-SNE visualization of \OurMODEL{} data with Arctic-Embed 2.0~\citep{Arctic_Embed}.}
\label{fig:midtraining_tsne}
\end{minipage}
\endgroup
\end{minipage}
\par\medskip

\subsection{SFT Data}
\label{app:sft_data}

Our SFT mixture comprises five domains from three public datasets. Nemotron-SFT-Agentic-v2~\citep{nvidia2026nemotronagenticv2} provides three complementary subsets: \emph{interactive agent} contains multi-turn customer-service conversations involving tool use; \emph{tool calling} covers synthetic single-turn, multi-turn, and multi-step tool-use trajectories; and \emph{search} contains web-search trajectories for answering factual questions.

Nemotron-SFT-SWE-v3.5~\citep{nvidia2026nemotronswe35} provides software-engineering trajectories collected with OpenCode, covering changes to code, tests, documentation, and configuration files. Following the setting of ProLong~\citep{ProLong}, for general conversational instruction following, we use the \texttt{train\_sft} split of UltraChat-200K,\footnote{\href{https://huggingface.co/datasets/HuggingFaceH4/ultrachat_200k}{Hugging Face H4 UltraChat-200K dataset card.}} a filtered version of UltraChat~\citep{ding2023enhancing}.

\begin{table}[htbp]
\centering
\small
\setlength{\tabcolsep}{4pt}
\renewcommand{\arraystretch}{1.15}
\begin{tabular*}{\linewidth}{@{}l@{\extracolsep{\fill}}rrrr@{}}
\toprule
\textbf{Domain} & \textbf{Records} & \textbf{Tokens (B)} & \textbf{Sampling (\%)} & \textbf{Token mix (\%)} \\
\midrule
Nemotron Interactive & 278,880 & 1.220 & 37.83 & 40.01 \\
Nemotron Tool Calling & 707,052 & 3.804 & 36.94 & 36.86 \\
Nemotron SWE & 5,115 & 0.309 & 12.47 & 10.07 \\
UltraChat & 207,865 & 0.247 & 7.58 & 8.08 \\
Nemotron Search & 5,968 & 0.152 & 5.18 & 4.97 \\
\midrule
\textbf{Total} & \textbf{1,204,880} & \textbf{5.732} & \textbf{100.00} & \textbf{100.00} \\
\bottomrule
\end{tabular*}
\caption{SFT corpus statistics and mixture proportions.}
\label{tab:sft_data_statistics}
\end{table}

Table~\ref{tab:sft_data_statistics} reports corpus sizes before mixture sampling. Records are complete conversations, and token counts include the Qwen chat template before overlength splitting and packing, without extra BOS/EOS tokens. Sampling weights apply to packed sequences of up to 131,072 tokens. The expected training-token share for domain $i$ is $w_i\bar{L}_i/\sum_j w_j\bar{L}_j$, where $w_i$ is its sampling weight and $\bar{L}_i$ its mean packed length. Supervision covers rendered assistant reasoning, responses, and tool calls, while system messages, user messages, and tool observations are masked.

\subsection{Benchmark Decontamination}
\label{app:benchmark_decontamination}

We screen the raw agent trajectories, generated question--answer (QA) examples, approximately 30B-token background corpus, and downstream SFT data against benchmark test data. Screening combines $n$-gram matching with semantic matching to identify lexical overlap and semantically equivalent content, including rephrased test material. Both source trajectories and compiled QA examples are checked, covering the original interactions and their derived supervision.

For trajectory-derived data, exclusion operates at the source-trajectory level. If a trajectory or any of its generated QA examples is flagged, we discard the entire trajectory together with all QA examples derived from it, rather than removing only the matched passage or individual QA pair. Matched records in the background corpus and SFT data are likewise excluded from their respective training mixtures.

\FloatBarrier
\section{Benchmarks and Baselines}
\label{app:benchmarks_baselines}

This section describes the baseline models, followed by the long-context and agent benchmarks used for downstream SFT and mid-training ICL evaluation.

\subsection{Baseline Models}

\paragraph{GPT-OSS-120B.}
OpenAI's gpt-oss-120b~\citep{agarwal2025gpt} is an open-weight mixture-of-experts reasoning model with approximately 117B total and 5.1B active parameters per token. It supports configurable reasoning effort and native tool calling.

\paragraph{Gemini 2.5 Flash.}
Gemini 2.5 Flash~\citep{comanici2025gemini} is Google's multimodal model designed to balance reasoning quality, latency, and computational cost. It supports a controllable thinking budget, long-context input, and tool use; our tables report the thinking configuration.

\paragraph{QwenLong-L1.5-30B-A3B.}
QwenLong-L1.5~\citep{DBLP:journals/corr/abs-2512-12967} builds on Qwen3-30B-A3B-Thinking, combining synthesized long-context reasoning tasks with specialized reinforcement learning. Its reported 256K context window is distinct from the memory-management framework used for longer inputs.

\paragraph{Qwen3-14B and Qwen3-30B-A3B.}
Both models belong to the Qwen3 family~\citep{yang2025qwen3} and support thinking and non-thinking modes. Qwen3-14B is a dense model with 14.8B parameters, whereas Qwen3-30B-A3B uses a mixture-of-experts architecture with 30.5B total and 3.3B active parameters. The released models have a native context length of 32,768 tokens; our baselines use inference-time context extension with YaRN~\citep{YaRN}.

\paragraph{ACC (SFT).}
We use the publicly released ACC dataset~\citep{su2026acc} for the baseline in Table~\ref{tab:training_stage_comparison}. We mix ACC with the same original SFT corpus used by \OurMODEL{} (SFT) at a 3:7 ratio and fine-tune Qwen3-14B-Base for two effective epochs. The total input-token budget and all remaining SFT settings are identical to those of \OurMODEL{} (SFT), as detailed in Appendix~\ref{app:training_details}.

\subsection{Downstream Long-context SFT Benchmarks}
\label{app:long_context_evaluation}

For downstream SFT evaluation, we strictly follow the official evaluation protocols of the following six public long-context benchmarks.

\paragraph{LongBench v2.}
LongBench v2~\citep{bai2025longbench} evaluates long-context understanding and reasoning through multiple-choice questions covering single- and multi-document QA, in-context learning, dialogue history, code repositories, and structured data.

\paragraph{MRCR.}
Multi-round Co-reference Resolution (MRCR), described in Michelangelo~\citep{vodrahalli2024michelangelo}, requires locating and reproducing a specified response among repeated, similar requests in a long conversation. The OpenAI release~\citep{openai2025mrcr} includes two-, four-, and eight-needle settings, testing retrieval with order-sensitive disambiguation. We compute scores separately for the 2-, 4-, and 8-needle settings and report their unweighted mean.

\paragraph{GraphWalks.}
GraphWalks~\citep{openai2025graphwalks} presents a directed graph as an edge list and asks for nodes satisfying a breadth-first-search or parent-retrieval query, testing dependency tracing over long structured inputs. We compute scores separately for the 0--128K and 128--256K context-length intervals and report their unweighted mean.

\paragraph{Oolong-Synth.}
The synthetic component of Oolong~\citep{bertsch2025oolong} requires interpreting individual text records and aggregating their properties to answer counting, comparison, and distributional questions across the context. Results are reported in Table~\ref{tab:oolong_synth}.

\paragraph{GSM-Infinite (GSM-Inf).}
GSM-Infinite~\citep{zhou2025gsm} generates grade-school arithmetic problems from computational graphs, allowing reasoning complexity and distractor-driven context length to be controlled separately.

\paragraph{CorpusQA.}
CorpusQA~\citep{lu2026corpusqa} evaluates corpus-level analysis that integrates evidence across many documents, including comparison and statistical aggregation, using synthesized questions with programmatically verifiable answers.

\subsection{Downstream Agent SFT Benchmarks}
\label{app:agent_evaluation}

We evaluate downstream SFT models on the following twelve public agent benchmarks, retaining their benchmark-specific scoring rules. The non-binary scores and thresholding conventions used in our tables are specified below; repeated-trial aggregation is detailed in Appendix~\ref{app:sft_evaluation_protocols}.

\paragraph{BrowseComp (BC).}
BrowseComp (Browsing Competition)~\citep{wei2025browsecomp} tests persistent web search for hard-to-find facts with short, verifiable answers. Its full evaluation set is denoted BC-full.

\paragraph{BrowseComp-ZH (BC-ZH).}
BrowseComp-ZH~\citep{zhou2025browsecomp} targets multi-hop information seeking on the Chinese web.

\paragraph{WideSearch.}
WideSearch~\citep{DBLP:journals/corr/abs-2508-07999} requires collecting numerous verifiable facts and organizing them into a complete structured answer. We report item-level F1, rather than exact-match accuracy. Avg@3 reports average item-level F1, while BoN@3 reports best-of-three item-level F1 (Item F1 Max@3).

\paragraph{FinSearchComp (FSC-V2).}
FinSearchComp~\citep{hu2026finsearchcomp} evaluates financial information retrieval and reasoning through time-sensitive data fetching, historical fact lookup, and complex historical investigations.

\paragraph{SealQA.}
SealQA~\citep{pham2026sealqa} tests fact-seeking under noisy, conflicting, or unhelpful search results.

\paragraph{MCP-Atlas.}
MCP-Atlas~\citep{bandi2026mcp} evaluates discovering and composing tools across real Model Context Protocol (MCP) servers to complete multi-step tasks. Gemini 2.5 Pro scores each reference claim as fulfilled, partially fulfilled, or unfulfilled, assigning 1, 0.5, or 0, respectively; their mean gives the response's claim coverage $c$. Avg@3 reports the mean binary success indicator $\mathbf{1}[c\geq 0.75]$, whereas BoN@3 averages the maximum \emph{unthresholded coverage} over three trials. These columns therefore measure thresholded pass rate and best-of-three coverage, not the mean and maximum of the same trial-level score.

\paragraph{SkillsBench.}
SkillsBench~\citep{li2026skillsbench} evaluates agent performance on diverse tasks paired with reusable procedural Skills and executable verifiers. 

\paragraph{WebArena.}
WebArena~\citep{zhou2024webarena} evaluates instruction-driven, multi-step interaction with reproducible websites, with success determined by the resulting task state.

\paragraph{Terminal-Bench 2.1 (TB 2.1).}
Terminal-Bench 2.1~\citep{merrill2026terminal,terminalbench2026v21} evaluates completing realistic tasks in executable terminal environments. Version 2.1 revises the task specifications, dependencies, and resource settings of Terminal-Bench 2.0.

\paragraph{NL2Repo-Bench (NL2Repo).}
NL2Repo-Bench~\citep{ding2025nl2repo} requires constructing an installable, multi-module Python library from a natural-language specification and an initially empty workspace. Its execution-based score is the fraction of reference tests passed, capped at 1. We average these normalized per-repository scores for Avg@3 and take their three-trial maximum for BoN@3, rather than requiring every test to pass for a repository to receive credit.

\paragraph{SWE-bench Verified (SWE-Verified).}
SWE-bench Verified~\citep{SWE_bench,openai2024sweverified} is a human-validated subset of SWE-bench: agents modify an existing repository to resolve a GitHub issue, with correctness checked by tests.

\paragraph{SWE-bench Pro (SWE-Pro).}
SWE-bench Pro~\citep{deng2025swe} extends repository-level issue resolution to longer-horizon, complex software-engineering problems across diverse projects.

\subsection{Mid-training Long ICL Benchmarks}
\label{app:long_icl_benchmarks}

Long ICL evaluates long-context capabilities directly at mid-training checkpoints, before downstream SFT. Our analyses cover ten tasks from five benchmark families: MRCR, RULER, LongBench v2, GraphWalks, and BrowseCompLong. Table~\ref{tab:long_icl_benchmarks} summarizes their tasks, output formats, and evaluation settings. For instruction-oriented benchmarks, we use completion prompts with task-specific demonstrations and answer cues to reduce dependence on chat-style instruction following. The adaptations below specify how the model should continue the prompt while retaining the underlying long-context task.

\begin{table}[!htbp]
    \centering
    \footnotesize
    \setlength{\tabcolsep}{3pt}
    \renewcommand{\arraystretch}{1.2}
    \begin{tabular}{@{}>{\raggedright\arraybackslash}p{0.20\linewidth}>{\raggedright\arraybackslash}p{0.13\linewidth}>{\raggedright\arraybackslash}p{\dimexpr0.39\linewidth-30pt\relax}>{\centering\arraybackslash}p{0.08\linewidth}>{\centering\arraybackslash}p{0.12\linewidth}>{\raggedleft\arraybackslash}p{0.08\linewidth}@{}}
        \toprule
        \textbf{Benchmark} & \textbf{Task} & \textbf{Description} & \textbf{Format} & \textbf{Length setting} & \textbf{Size} \\
        \midrule
        MRCR-ICL & Co-reference resolution & Reproduce an ordinally specified response among repeated requests, with the required prefix. & TC & $\leq$256K & 1,711 \\
        \midrule
        \multirow[t]{6}{*}{RULER-ICL} & RULER-MultiKey & Retrieve the queried key's value among key--value pairs and distractors. & TC & 4K--128K & 300 \\
        \cmidrule(l){2-6}
        & Multivalue & Retrieve all values associated with the queried key. & TC & 4K--128K & 300 \\
        \cmidrule(l){2-6}
        & QA1 & Answer a SQuAD question using a relevant passage among distractors. & TC & 4K--128K & 300 \\
        \cmidrule(l){2-6}
        & QA2 & Combine supporting documents to answer a HotpotQA question. & TC & 4K--128K & 300 \\
        \cmidrule(l){2-6}
        & CWE & Identify the 10 most frequent words in a long numbered list. & TC & 4K--128K & 300 \\
        \cmidrule(l){2-6}
        & VT & Follow assignment chains to return all variables sharing a queried value. & TC & 4K--128K & 350 \\
        \midrule
        LongBench v2-ICL & Long-context reasoning & Select the correct answer by reasoning over long documents, dialogue, code, or structured data. & MCQ & S/M/L & 503 \\
        \midrule
        GraphWalks-ICL & Directed graph queries & Return nodes at exactly a specified BFS depth or direct parents of a node in a directed graph. & TC & 4K--128K & 637 \\
        \midrule
        BrowseCompLong-ICL & Multi-document QA & Answer multi-hop questions by integrating supplied webpage evidence without external browsing. & TC & 20K--243K & 310 \\
        \bottomrule
    \end{tabular}
    \caption{Long ICL benchmark overview. MCQ denotes multiple-choice question answering; TC denotes text completion. Length denotes nominal settings or subset labels, except for BrowseCompLong-ICL, which reports the observed complete-prompt token range; S/M/L denotes short, medium, and long subsets. Size counts evaluated queries, excluding ICL demonstrations. MRCR includes the 2-, 4-, and 8-needle settings. RULER covers six selected subtasks, each evaluated at 4K, 8K, 16K, 32K, 64K, and 128K. BrowseCompLong-ICL uses three randomly selected demonstrations; its complete prompts span approximately 20K--243K tokens in the verified run.}
    \label{tab:long_icl_benchmarks}
\end{table}

\paragraph{MRCR-ICL.}
MRCR~\citep{openai2025mrcr} presents long conversations containing repeated requests with similar topics and formats. The final query identifies one earlier response by its occurrence order and asks the model to reproduce it with a specified random prefix. This design tests precise retrieval and disambiguation rather than topical matching. We use the 2-, 4-, and 8-needle settings with contexts up to 256K. Scoring first checks the required prefix; a missing or incorrect prefix receives zero. Otherwise, the prefix is removed and the remaining response is compared with the reference using \texttt{difflib.SequenceMatcher}.

\paragraph{RULER-ICL.}
We use six tasks from the RULER~\citep{hsieh2024ruler} evaluation configuration. \textit{RULER-MultiKey} (\texttt{niah\_multikey\_1}) retrieves the value of a queried key among competing key--value pairs, while \textit{Multivalue} retrieves all values associated with the same key. \textit{QA1} and \textit{QA2} place SQuAD~\citep{SQuAD} and HotpotQA~\citep{HotpotQA} evidence, respectively, among additional documents. \textit{Common Words Extraction} (CWE) requires identifying the 10 most frequent words in a long numbered list, and \textit{Variable Tracking} (VT) requires following assignment chains to identify all variables with a specified value. Their context--query--answer templates provide task-specific answer prefixes, supporting direct continuation by base models.

\paragraph{LongBench v2-ICL.}
We use the documents, questions, and four answer options from the \texttt{truncated\_by\_qwen\_100k} version of LongBench v2~\citep{bai2025longbench}. We use a two-shot format with synthetic short-document multiple-choice QA demonstrations. Each demonstration gives the selected option; the target prompt ends with \texttt{The correct answer is (} to elicit an answer letter. Predictions are evaluated against the reference option. The short, medium, and long subsets contain 180, 215, and 108 questions, respectively; these are source subset labels, not final input-length ranges.

\paragraph{GraphWalks-ICL.}
GraphWalks~\citep{openai2025graphwalks} requires returning the nodes at exactly a specified breadth-first-search depth or the direct parents of a queried node in a directed graph. We prepends four independent demonstrations: BFS at depths one and two, and parent queries with nonempty and empty answers. Each demonstration contains its own graph, operation, and answer, avoiding a pattern that encourages the model to generate another operation instead of answering. The test graph and operation follow, ending with \texttt{Final Answer: [} so generation continues inside the answer list. We remove test instances with empty gold node sets to prevent credit from trivial empty-list responses. The reported evaluation uses 637 questions across six length groups labeled 4K--128K.

\paragraph{BrowseCompLong-ICL.}
BrowseCompLong\footnote{\url{https://huggingface.co/datasets/openai/BrowseCompLongContext}} adapts BrowseComp~\citep{wei2025browsecomp} to multi-document QA, requiring evidence aggregation across supplied webpage documents without live browsing. Each prompt includes three randomly selected, answered demonstrations followed by a fourth, unanswered query. The 32K label refers to the base-dataset setting; complete evaluation prompts, including demonstrations, span 20,244–243,265 tokens. The model generates a short text answer, evaluated using \texttt{ConsistencyEval}. Responses are instructed to end with \texttt{Finished}, separated by two newlines on each side; when this delimiter is present, only the preceding text is retained for scoring.

\paragraph{Score aggregation.}
We average scores equally across the three MRCR needle settings, the six GraphWalks length groups, and the three LongBench v2 length subsets. For all six RULER tasks (RULER-MultiKey, Multivalue, QA1, QA2, CWE, and VT), reported scores equally average 4K, 8K, 16K, 32K, 64K, and 128K settings. The GraphWalks + LongBench v2 learning-dynamics proxy gives equal weight to the two benchmark means rather than pooling their individual settings. Gains use the same settings as their reference model, and are reported in percentage points.

\FloatBarrier
\subsection{Mid-training Agent ICL Benchmarks}
\label{app:agent_icl_benchmarks}

Following APTBench~\citep{qin2025aptbench}, we formulate agent tasks as multiple-choice or text-completion probes to assess base models' agent atomic capabilities before SFT. We adapt and extend this design with line-level bug localization and dedicated probes for code-summary selection, failure-aware multi-hop replanning, tool selection, and argument filling.

Agent ICL evaluates individual agent capabilities at mid-training checkpoints, before downstream SFT. We use seven tasks derived from public datasets, including SWE-bench Verified and Lite~\citep{SWE_bench,openai2024sweverified}, BFCL~\citep{patil2025berkeley}, TACO~\citep{li2023taco}, PlanBench-XL~\citep{liu2026planbench}, and MCP-Bench~\citep{wang2026mcp}. They cover code understanding and repair, planning and replanning, and tool use, complementing end-to-end agent evaluations. Task-specific completion prompts elicit a line number or an answer choice from the supplied task context, with demonstrations where applicable. All tasks report accuracy. Table~\ref{tab:agent_icl_benchmarks} summarizes their objectives, output formats, and sizes.

\begin{table}[!htbp]
    \centering
    \footnotesize
    \setlength{\tabcolsep}{3pt}
    \renewcommand{\arraystretch}{1.2}
    \begin{tabular}{@{}>{\raggedright\arraybackslash}p{0.25\linewidth}>{\raggedright\arraybackslash}p{\dimexpr0.57\linewidth-18pt\relax}>{\centering\arraybackslash}p{0.10\linewidth}>{\raggedleft\arraybackslash}p{0.08\linewidth}@{}}
        \toprule
        \textbf{Benchmark} & \textbf{Description} & \textbf{Format} & \textbf{Size} \\
        \midrule
        Bug Localization & Locate the faulty line using an issue description and numbered code. & TC & 114 \\
        \midrule
        Bug Fixing & Select the correct replacement for a specified faulty line. & MCQ & 114 \\
        \midrule
        Code Summary & Identify the summary that matches a code snippet's behavior. & MCQ & 235 \\
        \midrule
        Planning Preference & Choose the next action from the task goal and interaction history. & MCQ & 181 \\
        \midrule
        Multi-hop Replanning & Select a feasible continuation or task status after tool failure. & MCQ & 210 \\
        \midrule
        Tool Selection & Match a user request to a tool using candidate descriptions. & MCQ & 103 \\
        \midrule
        Argument Filling & Select a missing argument value using the interaction history. & MCQ & 325 \\
        \bottomrule
    \end{tabular}
    \caption{Agent ICL benchmark overview. MCQ denotes multiple-choice question answering; TC denotes text completion.}
    \label{tab:agent_icl_benchmarks}
\end{table}

\paragraph{Bug Localization.}
This task connects issue understanding with fault localization: the model identifies the line responsible for a reported defect in a numbered source file. Three labeled demonstrations establish the completion format, and the test prompt ends with \texttt{The buggy line number is:} to elicit a line number.

\paragraph{Bug Fixing.}
With the faulty line identified, the model must distinguish a correct repair from three alternative replacements using the issue description and surrounding code. Three demonstrations illustrate the four-choice format. The prompt ends with \texttt{Answer:}, where the four candidate letters are scored.

\paragraph{Code Summary.}
Code Summary tests whether a model understands a snippet's functionality and behavior by choosing among four candidate summaries. Differences in conditions, operation order, or returned quantities require attention to program semantics. The model completes an \texttt{Answer:} cue with the selected letter.

\paragraph{Planning Preference.}
The user query and preceding search or browsing history provide the evidence for choosing the next action from six candidates. This tests whether a model can identify productive follow-up actions and recognize when enough information has been gathered to conclude the task. Three demonstrations precede the test question, which ends with \texttt{Your choice:}.

\paragraph{Multi-hop Replanning.}
Tool unavailability can invalidate a previously feasible plan. Using tool descriptions, execution history, and failure feedback, the model chooses among four alternatives representing a continuation plan, a next action, or a task-status judgment. These questions probe dependency tracking and adaptation to disrupted execution paths.

\paragraph{Tool Selection.}
The model selects a tool from three or four candidates for a user request by interpreting the supplied function descriptions. Closely related tool functions and differing constraints test semantic discrimination beyond name matching. Three demonstrations illustrate the task before the model completes the test question's \texttt{Answer:} cue.

\paragraph{Argument Filling.}
This task focuses on binding information from an interaction history to the next tool call. A missing argument is marked in the call, and the model chooses its value from four candidates using the tool descriptions and previous observations. Correct choices require resolving entities and their roles in context.

\paragraph{Scoring and aggregation.}
Each question yields one scored prediction. Bug Localization, Code Summary, Planning Preference, and Tool Selection use greedy completion at temperature zero, followed by extraction of the predicted line number or answer letter. Bug Fixing, Multi-hop Replanning, and Argument Filling instead select the candidate letter with the highest final-token log probability at the answer position. Predictions are checked against the reference answers. Multi-hop Replanning uses overall question accuracy rather than a macro-average across its question categories.

Bug Localization and Bug Fixing each contain 82 Python and 32 Multilingual questions; we give the two subset accuracies equal weight. The Tool Selection + Argument Filling proxy in the learning-dynamics analysis likewise averages the two task scores equally. Reported gains use the reference model specified in each figure and are expressed in percentage points.

\FloatBarrier

\section{Evaluation Protocols}
\label{app:evaluation_protocols}

\subsection{ICL Evaluation}
\label{app:icl_evaluation_protocols}

\paragraph{Inference Settings.}
We evaluate mid-training checkpoints before SFT using Long ICL, Agent ICL, and general ICL, described in Appendices~\ref{app:long_icl_benchmarks}, \ref{app:agent_icl_benchmarks}, and~\ref{app:general_icl}. Agent ICL probes individual atomic capabilities from supplied contexts, without executing an interactive agent scaffold. We use a completion interface with temperature 0 and a default generation limit of 1,024 tokens. The serving context limits are 262,144 tokens for Long and Agent ICL and 65,536 tokens for general ICL. All ICL and SFT evaluations set \texttt{truncation\_side=right}. No separate thinking budget is specified for ICL.

\paragraph{Long ICL.}
For Long ICL, we generate one greedy completion per question, with the task-specific scoring and subset aggregation described in Appendix~\ref{app:long_icl_benchmarks}. Generation is capped at 1,024 tokens for MRCR, RULER, GraphWalks, LongBench v2, and BrowseCompLong. LongBench v2 additionally stops at a closing parenthesis or newline. We specify no custom stop sequence for the other four tasks.

\paragraph{Agent ICL.}
Agent ICL uses one trial per question. Bug Localization, Code Summary, Planning Preference, and Tool Selection use greedy completion; Bug Fixing, Replanning, and Argument Filling instead rank answer candidates using terminal-token log-probabilities. For these likelihood-based probes, requests with \texttt{max\_tokens=1} obtain prompt log-probabilities; candidate-wise requests are not repeated sampled trials.

\paragraph{General ICL.}
General ICL similarly retains benchmark-specific scoring: PIQA scores candidate continuations, MMMLU-Lite uses terminal-token log-probabilities, and AGIEval-CN generates an answer with a newline stop condition.

\subsection{SFT Evaluation}
\label{app:sft_evaluation_protocols}

\paragraph{Generation Settings.}
After SFT, we evaluate long-context question answering and interactive agent tasks separately from the ICL probes. For Qwen3-14B, we use task-level defaults of temperature 0.7, top-$p$ 1.0, no top-$k$ cutoff, and a repetition penalty of 1.0. The serving context window is 262,144 tokens, with a default output cap of 65,536 tokens per model call. We specify no custom stop sequence at the task level. Results are reported using Avg@3 and BoN@3, as defined below.

\paragraph{Coding and Tool-use Agents.}
We execute coding and tool-use agents with benchmark-specific scaffolds. For SWE-bench Verified and SWE-Pro, we use \texttt{swe\_agent}~\citep{SWE_agent} with at most 250 iterations and bounded retries for tool-error recovery. SWE-bench Verified additionally uses a 131,072-token context limit. For Terminal-Bench, we use \texttt{terminus2} with XML action parsing, at most 500 iterations, context summarization, and a 229,376-token context limit. NL2Repo uses the Claude Code scaffold with at most 250 turns. MCP-Atlas uses an MCP-enabled agent with at most 100 iterations. Agent execution and benchmark evaluation are subject to task-specific timeouts.

\paragraph{Search Agents.}
We use the same agent execution settings across all search benchmarks: web search, webpage fetching, and a Python interpreter, with at most 200 iterations and history summarization. Bounded retries handle failed model calls, invalid tool arguments, and tool-call errors. Retrieved webpage content is length-limited and progressively truncated when needed to fit the context budget.

\paragraph{Scoring and Repeated Trials.}
We retain benchmark-specific scoring: CorpusQA uses GPT-OSS-120B, and SWE-bench Verified uses patch execution with \texttt{FAIL\_TO\_PASS} and \texttt{PASS\_TO\_PASS} tests. Agent score definitions appear in Appendix~\ref{app:agent_evaluation}. In particular, MCP-Atlas Avg@3 averages thresholded success indicators, whereas its BoN@3 uses continuous claim coverage; WideSearch and NL2Repo retain graded scores in both Avg@3 and BoN@3. Failed trials receive zero and remain in the denominator, including timeouts, unrecovered execution or tool errors, invalid answers, and invalid scoring records.

We report average and best-of-$N$ performance at $N=3$ (Avg@3 and BoN@3), averaged across examples using the benchmark-specific scores above. For MRCR and GraphWalks, this averaging is performed within each subset before the equal-weight aggregation specified in Appendix~\ref{app:long_context_evaluation}. Binary scores measure task success, while graded scores retain partial credit.

\FloatBarrier
\section{Implementation Details}
\label{app:implementation_details}

\subsection{Training Details}
\label{app:training_details}

We adopt the open-source ProLong framework~\citep{ProLong} for mid-training and downstream supervised fine-tuning. Tables~\ref{tab:midtraining_configuration} and~\ref{tab:sft_configuration} report the mid-training and downstream SFT settings for Qwen3-14B and Qwen3-30B-A3B under a 256-GPU reference configuration. Both stages retain 262,144 position embeddings and use boundary-preserving packing. Mid-training supervises all valid prediction positions; SFT supervises assistant reasoning, responses, and tool calls with running-mean valid-token normalization.

\paragraph{Mixture Sampling and Token Accounting.}
Mixture proportions are defined by input-token counts. For the All variant, Retrieval, MultidocQA, and State Tracking contribute 3B, 4B, and 3B tokens, respectively. Each single-capability variant uses its own 10B-token corpus without repeated sampling. All token budgets refer to actual input tokens processed during training.

\begin{table}[!htbp]
    \centering
    \small
    \setlength{\tabcolsep}{5pt}
    \renewcommand{\arraystretch}{1.05}
    \begin{tabular*}{\linewidth}{@{}l@{\extracolsep{\fill}}l@{}}
        \toprule
        \textbf{Parameter} & \textbf{Configuration} \\
        \midrule
        Maximum sequence length & 262,144 tokens \\
        RoPE base & 16,000,000 (no additional scaling) \\
        Optimizer & AdamW \\
        Adam $\beta_1$, $\beta_2$; $\epsilon$ & $(0.9, 0.95)$; $10^{-8}$ \\
        Peak / minimum learning rate & $4\times10^{-5}$ / $4\times10^{-6}$ \\
        Learning-rate schedule / warmup & Cosine / 10\% \\
        Weight decay / gradient clipping & 0.1 / 1.0 \\
        \midrule
        Global / per-device microbatch size & 64 / 1 \\
        Reference GPU count & 256 \\
        Sequence-parallel degree & 8 \\
        Gradient accumulation & 2 \\
        Distributed training & FSDP full sharding + sequence parallelism \\
        Precision / random seed & BF16 mixed precision / 42 \\
        Gradient checkpointing & Enabled \\
        Checkpoint interval & Every 300 steps and the final step \\
        \bottomrule
    \end{tabular*}
    \caption{Mid-training configuration. Global batch size counts independent packed sequences, excluding sequence-parallel replicas.}
    \label{tab:midtraining_configuration}
\end{table}

\begin{table}[!htbp]
    \centering
    \small
    \setlength{\tabcolsep}{5pt}
    \renewcommand{\arraystretch}{1.05}
    \begin{tabular*}{\linewidth}{@{}l@{\extracolsep{\fill}}l@{}}
        \toprule
        \textbf{Parameter} & \textbf{Configuration} \\
        \midrule
        Maximum sequence length & 131,072 tokens \\
        RoPE base & 16,000,000 (no additional scaling) \\
        Optimizer & AdamW \\
        Adam $\beta_1$, $\beta_2$; $\epsilon$ & $(0.9, 0.95)$; $10^{-8}$ \\
        Peak / minimum learning rate & $2\times10^{-5}$ / $2\times10^{-6}$ \\
        Learning-rate schedule / warmup & Cosine / 3\% \\
        Weight decay / gradient clipping & 0.0 / 1.0 \\
        \midrule
        Global / per-device microbatch size & 32 / 1 \\
        Reference GPU count & 256 \\
        Sequence-parallel degree & 8 \\
        Gradient accumulation & 1 \\
        Distributed training & FSDP full sharding + sequence parallelism \\
        Precision / random seed & BF16 mixed precision / 42 \\
        Gradient checkpointing / attention & Enabled / FlashAttention-3 \\
        Logit block size & 2,048 \\
        \bottomrule
    \end{tabular*}
    \caption{Downstream SFT configuration. The positional configuration is retained from mid-training, while training sequences are capped at 128K tokens.}
    \label{tab:sft_configuration}
\end{table}

\FloatBarrier

\subsection{Retrieval Data Construction}
\label{app:retrieval_construction}

Retrieval converts recorded search passages and file contents into exact-reproduction questions over the complete preprocessed trajectory. We extract candidate items, construct questions that distinguish similar observations or repeated file interactions, align reference answers with the rendered context, and obtain and filter teacher-generated solutions.

\paragraph{Evidence Extraction and Processing.}
We preserve the original interaction order and associate each candidate with its trajectory position and source metadata. Candidate extraction differs between search and code trajectories:
\begin{itemize}[label={$\bullet$},leftmargin=1.2em,itemsep=2pt,topsep=3pt,parsep=0pt]
    \item \textbf{Search passages.} We unpack structured search responses into individual passages, accommodating passage, snippet, and content fields across tool formats. Results from the same search response form a candidate group in their original order. We retain the associated query, result titles, and passage positions. These groups provide topically related candidates without inserting synthetic distractors.
    \item \textbf{File contents.} Read targets are file contents returned in tool observations, whereas write targets are content payloads supplied in tool-call arguments, rather than write acknowledgments. We normalize tool names, recover readable text from structured payloads, and remove system reminders. We match calls and responses only when their correspondence is unambiguous, recording the file path, operation type, and occurrence for each candidate.
\end{itemize}

\paragraph{Question Construction.}
For search trajectories, an LLM summarizes each group's query and result titles into a short topic phrase, falling back to the original query when needed. Eligible groups contain at least five passages; target passages must contain at least 100 characters and occur only once among the trajectory's extracted search passages. We deduplicate candidate groups by topic and sample one eligible target per selected group. When more than 12 groups remain, we retain the 12 with the largest total eligible passage lengths. Questions specify the topic and the target's original one-indexed position within its group. Seventy percent of search questions additionally require a random 10-character alphanumeric prefix before the reproduced passage, while the remaining 30\% request the passage alone. We do not renumber passages after target filtering or add extra questions from an already selected group to fill the question budget.

For code trajectories, questions specify the file path, read/write operation, and occurrence, distinguishing repeated interactions with the same file. We sample read and write targets at a target ratio of 7:3, filling shortages from the other type. Requests may additionally require a random prefix before the reproduced content; 30\% of code questions use this constraint. Both prefixed and plain requests ask for the recorded content, not a summary or a reconstructed file state.

\paragraph{Context Rendering and Reference Alignment.}
We render the complete preprocessed trajectory without further truncating tool observations. Each search passage receives its group's topic header, while file blocks retain explicit read/write cues. Reference answers are recovered from the rendered target, rather than copied from an earlier raw representation, so that formatting cleanup and consistent de-identification do not create answer--context mismatches. We require the rendered target to occur exactly once in the context and discard questions with missing or ambiguous alignments. Other trajectory content remains available as context and distractors.

\paragraph{Solution Generation and Filtering.}
We use DeepSeek-v4-pro~\citep{DeepSeek_V4} as the teacher model to solve retrieval questions given the rendered trajectory. We compare the final answer extracted from each solution with the reference answer, checking required prefixes and applying the same fidelity criteria to search and code questions. Exact matches and matches under Unicode and whitespace normalization are accepted directly. Otherwise, solutions are retained only when their final answers achieve a \texttt{SequenceMatcher} similarity of at least 0.95 with the reference. Lower-scoring solutions and failed generation requests are discarded. We append accepted question--solution pairs after the shared trajectory context and remove records with no surviving questions. Table~\ref{tab:case_retrieval} gives an example.

\Needspace{10\baselineskip}
\subsection{MultidocQA Data Construction}
\label{app:multidocqa_construction}

MultidocQA converts the preprocessed trajectories into questions that require
combining evidence across sources. We first extract and organize the evidence,
then construct and refine questions, and finally obtain teacher-generated
solutions using the complete evidence collection.

\paragraph{Evidence Extraction and Processing.}
We split each trajectory into interaction turns initiated by user messages.
GPT-OSS-120B groups these turns according to continuity of the task, the objects
being manipulated, and the resulting artifacts. Turns that develop the same
feature or research deliverable, including follow-up requests and related
debugging or verification, belong to the same group. A group may contain
nonconsecutive turns when the user returns to an earlier task. We check that
the groups cover all turns without duplication.

Within each group, we extract file, document, and execution evidence from tool
interactions. We check content validity and exclude empty or uninformative
responses, inaccessible or removed content, binary data, visibly corrupted
text, and fragments below minimum length thresholds. We then process each
evidence type as follows:
\begin{itemize}[label={$\bullet$},leftmargin=1.2em,itemsep=2pt,topsep=3pt,parsep=0pt]
    \item \textbf{File evidence.} We extract code and configuration contents
    from file reads and write/edit operations. We remove system reminders,
    terminal escape sequences, and non-printing control characters, retaining
    the recorded code/configuration text and edit fragments.
    \item \textbf{Document evidence.} We collect retrieved passages, fetched
    webpage text, and textual documents read or produced during the task.
    We unpack structured responses to recover titles and text, removing tool
    wrappers, image markup, and URL or metadata clutter.
    \item \textbf{Execution evidence.} We extract standard output and error
    text from command results, remove runtime-injected noise, and filter
    failure outputs without useful content. Previously observed contents of
    referenced scripts or files supplement the results when available.
\end{itemize}

We retain groups with at least four distinct source keys, based on file paths,
commands, or document content; versions of the same file count as one source.
Each item receives a unique identifier and source metadata. For each question
$q_j$, the generator selects substantive, interpretable supporting evidence
$S_j\subseteq E(\tau)$ from the retained evidence collection $E(\tau)$.

\paragraph{Version Organization and Context Construction.}
Observed contents of the same file share a source identity, with the version
label advancing when the content changes. To construct the long context, we
group evidence by source, order available versions within each group, and
randomly permute the source groups. Each block displays its evidence identifier,
source title, and version label when available. Temporal questions use explicit
version or time cues rather than block positions.

\paragraph{Question Construction.}
The process comprises three stages: diverse question generation, enrichment
of evidence dependencies, and final refinement and checking.

\emph{Stage 1: Diverse question generation.}
Gemini-3.5-Flash generates questions, reference answers, and supporting evidence
identifiers from each annotated task group. We use nine question categories and
nine answer categories (Table~\ref{tab:multidocqa_categories}). For each group,
we sample two question categories uniformly and two answer categories from the
union of compatible answer categories using category-specific weights; both
samplings are without replacement. Answer sampling favors lists, multi-point
analyses, and constrained outputs. Each question must use a compatible category
pairing, be answerable from the evidence independently of the agent dialogue,
and omit evidence identifiers. We validate supporting identifiers and require
at least two distinct sources.

\emph{Stage 2: Enriching multi-hop dependencies.}
We strengthen cross-source dependencies through up to two refinement rounds.
Each round supplies the current supporting set and up to ten additional
candidates, combining relevance ranking by shared entities, version relations,
and tool-interaction proximity with random selection. The model is prompted to
extend the question by linking facts into a multi-hop chain or combining
complementary constraints. Added evidence should contribute to the solution,
while the question category and answer requirements remain unchanged.

\emph{Stage 3: Final refinement and checking.}
The model prunes redundant supporting items from $S_j$, while retaining
$E(\tau)$ for the long context, and reviews answer support, ambiguity, and
distractor validity. It clarifies wording and output requirements and removes
solution steps or explicit retrieval routes from questions. Rule-based checks
validate evidence identifiers, category compatibility, and option--answer
consistency; semantic concerns are recorded for subsequent solution checks.

\begin{table}[htbp]
    \centering
    \small
    \setlength{\tabcolsep}{5pt}
    \renewcommand{\arraystretch}{1.12}
    \begin{tabular}{@{}>{\raggedright\arraybackslash}p{0.32\linewidth}>{\raggedright\arraybackslash}p{\dimexpr0.68\linewidth-10pt\relax}@{}}
        \toprule
        \textbf{Question category} & \textbf{Reasoning requirement} \\
        \midrule
        Cross-document or version comparison & Identify shared conclusions, differences, or changes across related documents or versions. \\
        Multi-hop fact composition & Use information from one source to locate and interpret the next source in a chain. \\
        Quantitative reasoning & Combine numerical facts to compute a derived quantity with a meaningful interpretation. \\
        Aggregation & Count, rank, deduplicate, or identify extrema across related records. \\
        Relational reasoning & Connect evidence through calls, data flow, or relationships between definitions and their use. \\
        Conditional reasoning & Infer an outcome under stated conditions or a hypothetical change. \\
        Global synthesis & Integrate the main conclusions, positions, or design rationale across sources. \\
        Temporal or causal ordering & Establish event order or causal relations using explicit cues in the evidence. \\
        Fine-grained discrimination & Distinguish supported or unsupported statements among closely related alternatives. \\
        \midrule
        \textbf{Answer category} & \textbf{Output requirement} \\
        \midrule
        One-sentence answer & A single natural-language sentence stating the requested conclusion. \\
        Entity & A specific name, identifier, path, term, or other short factual item. \\
        Numerical answer & A numerical value with specified units or precision. \\
        Single choice & The label of one selected option. \\
        Multiple choice & The labels of all selected options. \\
        List & A collection or sequence of items, with ordering or fields specified when needed. \\
        Multi-point analysis & An explanation covering the requested aspects of the answer. \\
        Constrained output & An answer satisfying explicit structural or formatting requirements, such as a specified JSON schema. \\
        Verbatim quotation & Exact reproduction of a requested evidence span. \\
        \bottomrule
    \end{tabular}
    \caption{Question and answer categories used to construct MultidocQA data.
    Question categories guide the reasoning task, while answer categories
    specify response requirements. Only compatible combinations are used.}
    \label{tab:multidocqa_categories}
\end{table}

\paragraph{Solution Generation.}
For each question, DeepSeek-v4-pro receives the complete retained evidence
collection $E(\tau)$ and that question, without its reference answer or
annotated supporting set. We encourage it to use evidence identifiers to
track relevant information while solving the question. All questions from a trajectory
share the same evidence order and are solved independently. After checking
final-answer agreement and solution consistency, we combine accepted questions
into a shared-context training record. For each question, the training response
combines the teacher-generated solution steps with the reference answer.
Table~\ref{tab:case_multidocqa} gives an example.

\FloatBarrier
\subsection{State Tracking Data Construction}
\label{app:state_tracking_construction}

State Tracking converts coding-agent trajectories into questions about evolving file states and the origins of their contents. We reconstruct per-file histories, derive reference answers programmatically, and obtain teacher-generated solutions for two question families: \emph{Snapshot} and \emph{Provenance}.

\paragraph{Evidence Extraction and Processing.}
We deduplicate trajectories within each agent harness using normalized user requests and tool/file-event sequences, retaining more complete representatives when one trajectory is a prefix of another. Tool calls are aligned with their responses using call identifiers. We normalize file paths and operation names and retain message positions, operation arguments, and execution outcomes. Events are then organized into per-file timelines. Only explicitly successful operations contribute to state updates; failed calls are not counted as successful modifications, and uncertain execution outcomes are not treated as confirmed updates.

\paragraph{State Reconstruction and Verification.}
A reconstruction segment begins from a complete Write payload or a Read whose completeness can be established from its arguments and returned contents. Partial or truncated reads are not treated as complete initial states. We replay successful edits in order, checking that each replacement can be applied unambiguously and preserving untouched content. A Write replaces the complete file state. When available, subsequent complete reads are compared with the reconstructed state. We distinguish read-verified histories from deterministic histories supported only by complete Write or strict Read anchors, rather than requiring a later read for every sample. Unreplayable edits, uncertain execution outcomes, ambiguous same-file ordering, or inconsistent observations break the segment. Reconstruction can resume from a new complete anchor. Potential unobserved changes from shell commands or sub-agents are recorded as separate risk indicators.

\paragraph{Question Construction.}
The two families share the reconstructed histories but query different aspects of state evolution. Table~\ref{tab:state_tracking_categories} summarizes their seven question subtypes and corresponding answer formats:
\begin{itemize}[label={$\bullet$},leftmargin=1.2em,itemsep=2pt,topsep=3pt,parsep=0pt]
    \item \textbf{Snapshot.} Questions request the complete file immediately before or after a specified successful Edit, at the end of the retained history, or after rolling back a suffix of consecutive edits. Reference answers are the corresponding reconstructed snapshots. Rollback is restricted to the supported edit suffix rather than arbitrary deletion of an earlier operation. These questions require composing partial updates while preserving the rest of the file, not reproducing an individual edit payload.
    \item \textbf{Provenance.} Questions identify the last operation that wrote a surviving fragment, the operation after which it first appeared, or the operation that removed it. During replay, content spans retain their source labels unless overwritten, while a full Write resets their provenance. Candidate fragments must have an unambiguous answer. Removal questions additionally require the fragment to exist immediately before the selected operation, disappear immediately afterward, and remain absent from all subsequent known states. Answers specify an operation type and its successful-operation index within the target file. Edit, Write, and Delete are counted separately from one, while an initial anchor uses index zero.
\end{itemize}
We diversify question wording using Chinese and English template pools and LLM-based paraphrasing, while keeping the target file, temporal boundary, operation-counting convention, and output requirements explicit and unchanged.

\begin{table}[htbp]
    \centering
    \small
    \setlength{\tabcolsep}{5pt}
    \renewcommand{\arraystretch}{1.12}
    \begin{tabular}{@{}>{\raggedright\arraybackslash}p{0.32\linewidth}>{\raggedright\arraybackslash}p{\dimexpr0.68\linewidth-10pt\relax}@{}}
        \toprule
        \textbf{Snapshot subtype} & \textbf{Question target} \\
        \midrule
        After edit & Recover the complete file immediately after its $K$-th successful Edit. \\
        Before edit & Recover the complete file immediately before its $K$-th successful Edit, incorporating preceding updates. \\
        Final state & Recover the complete file at the end of its reconstructible state history. \\
        Suffix rollback & Recover the file after undoing the last $R$ successful Edits in a consecutive suffix, without crossing a Write or Delete boundary. \\
        \midrule
        \textbf{Provenance subtype} & \textbf{Question target} \\
        \midrule
        Final-state last writer & Identify the Edit or Write that last wrote a target fragment still present in the final state. \\
        First introduction & Identify the earliest operation after which a target fragment appeared, or the initial anchor if it was already present. \\
        Removal & Identify the Edit, Write, or Delete that removes a previously present fragment, after which it remains absent from all subsequent known states. \\
        \midrule
        \textbf{Question family} & \textbf{Answer format} \\
        \midrule
        Snapshot & The complete target file contents, preserving unchanged text rather than returning an edit fragment or summary. \\
        Provenance & A structured answer specifying the operation type and its successful-operation index within the target file. Indices are counted separately by operation type from one; initial-anchor answers use zero when applicable. \\
        \bottomrule
    \end{tabular}
    \caption{Question subtypes and answer formats used to construct State Tracking data.}
    \label{tab:state_tracking_categories}
\end{table}

\paragraph{Context Construction and QA Organization.}
The rendered trajectory provides the context, and constructed questions are kept separate from the recorded agent dialogue. For Snapshot questions, the context ends before the first subsequent successful Read of the target file, including partial reads, to prevent direct copying of the requested state. If no such read exists, we retain the complete trajectory. Questions can share a training record only when their rendered contexts agree. Among Snapshot questions targeting the same state, we retain one formulation, with weighted sampling favoring final-state questions. Within each record, Snapshot questions follow the temporal order of their target states, and Provenance questions precede Snapshot questions about the same file. Before serialization, we additionally check cross-QA compatibility under this ordering. A QA combination is retained only if later answers cannot be directly inferred from complete file states supplied by preceding Snapshot answers, including for questions concerning the same file.

\paragraph{Difficulty Characterization.}
We characterize difficulty at the question level using successful target-file mutations before the dependency boundary, the queried operation index, context length, and interference from other files. Snapshot questions additionally track rollback depth and answer length; Provenance questions track mutations after the source event. Output length is treated separately from the complexity of reconstructing the state. Reconstruction confidence and hidden-mutation risk are quality attributes rather than difficulty labels.

\paragraph{Solution Generation and Filtering.}
We use DeepSeek-v4-pro as the teacher model to solve both Snapshot and Provenance questions using the rendered trajectory context. Snapshot final answers must match the reference under normalization and achieve a character-similarity score of at least 0.99. Provenance final answers must match the structured reference answer. We retain accepted teacher-generated solutions, including their original final answers. Each training record contains the shared trajectory context once, followed by the accepted question--solution pairs in their prescribed order. Table~\ref{tab:case_state_tracking} gives an example.

\FloatBarrier

\section{Additional Experiments and Analysis}
\label{app:additional_experiments}

\subsection{Oolong-Synth Results}
\label{app:oolong_synth}

Table~\ref{tab:oolong_synth} reports supplementary long-context information aggregation results under the same evaluation protocol as Table~\ref{tab:main_results}.

\begin{table*}[htbp]
\centering
\setlength\tabcolsep{7pt}
\renewcommand{\arraystretch}{1.45}
\fontsize{8.5pt}{9.5pt}\selectfont
\begin{tabular}{l|cc}
\toprule
\multicolumn{1}{c|}{\multirow{2}{*}{\textbf{Models}}}
& \multicolumn{2}{c}{\textbf{Oolong-Synth}} \\
\cmidrule(lr){2-3}
& Avg@3 & BoN@3 \\
\midrule
\multicolumn{3}{c}{\textbf{\textit{Frontier Models}}} \\
\midrule

Gemini-2.5-Flash-Thinking
& 78.17 & 91.00 \\

QwenLong L1.5-30B-A3B
& 75.50 & 89.00 \\

GPT-OSS-120B
& 68.00 & 84.50 \\

\midrule
\multicolumn{3}{c}{\textbf{\textit{Our Models}}} \\
\midrule

Qwen3-14B
& 52.17\zerodiff & 70.00\zerodiff \\

\hspace{0.4em}- SFT \textit{(w/o Mid-training)}
& 53.17\posdiff{1.0} & 71.00\posdiff{1.0} \\

\rowcolor{TrajRetrieval!12}
\hspace{0.4em}- \OurMODEL\ \textit{(Retrieval)}
& 57.17\posdiff{5.0} & 77.00\posdiff{7.0} \\

\rowcolor{TrajMultidocQA!12}
\hspace{0.4em}- \OurMODEL\ \textit{(MultidocQA)}
& 57.33\posdiff{5.2} & \textbf{79.00}\posdiff{9.0} \\

\rowcolor{TrajStateTracking!12}
\hspace{0.4em}- \OurMODEL\ \textit{(State Tracking)}
& 56.00\posdiff{3.8} & 75.50\posdiff{5.5} \\

\rowcolor{TrajAll!16}
\textbf{\OurMODEL-14B-All}
& \textbf{58.50}\posdiff{6.3} & 78.00\posdiff{8.0} \\

\midrule
Qwen3-30B-A3B
& 41.33\zerodiff & 52.50\zerodiff \\

\hspace{0.4em}- SFT \textit{(w/o Mid-training)}
& 52.33\posdiff{11.0} & \textbf{75.50}\posdiff{23.0} \\

\rowcolor{TrajAll!16}
\textbf{\OurMODEL-30B-A3B-All}
& \textbf{53.83}\posdiff{12.5} & 72.50\posdiff{20.0} \\
\bottomrule
\end{tabular}
\caption{SFT performance on Oolong-Synth. Subscripts show gains over the corresponding Qwen3 baseline (pp). The best scores for each backbone LLM are \textbf{bolded}.}
\label{tab:oolong_synth}
\end{table*}

\FloatBarrier
\Needspace{0.45\textheight}
\subsection{Additional Agent Evaluation Results}
\label{app:main_results_agent}

\begin{table*}[!htbp]
\centering
\setlength\tabcolsep{2.0pt}
\renewcommand{\arraystretch}{1.45}
\fontsize{8.5pt}{9.5pt}\selectfont

\resizebox{\linewidth}{!}{
\begin{tabular}{l|ccccc|ccc|cccc}
\toprule
\multicolumn{1}{c|}{\multirow{2}{*}{\textbf{Models}}}
& \multicolumn{5}{c|}{\textbf{Search}}
& \multicolumn{3}{c|}{\textbf{Tool Use}}
& \multicolumn{4}{c}{\textbf{Coding}} \\
\cmidrule(lr){2-6}
\cmidrule(lr){7-9}
\cmidrule(lr){10-13}

& \textbf{BC}
& \textbf{BC-ZH}
& \textbf{WideSearch}
& \textbf{FSC-V2}
& \textbf{SealQA}
& \textbf{MCP-Atlas}
& \textbf{SkillsBench}
& \textbf{WebArena}
& \textbf{TB 2.1}
& \textbf{NL2Repo}
& \textbf{SWE-Verified}
& \textbf{SWE-Pro} \\
\midrule

\multicolumn{13}{c}{\textbf{\textit{Frontier Models}}} \\
\midrule
\shortstack[l]{Gemini-2.5-Flash-Thinking}
& 12.72 & 43.42 & 44.09 & 46.44 & 38.75 & 47.47 & 16.67 & 51.98 & 34.83 & 12.40 & 48.60 & 24.08 \\

QwenLong L1.5-30B-A3B
& 6.48 & 21.35 & 45.14 & 32.51 & 28.02 & 62.77 & 4.17 & 36.12 & 4.49 & 15.09 & 43.60 & 19.56 \\

GPT-OSS-120B
& 37.91 & 38.41 & 51.01 & 55.75 & 34.23 & 69.83 & - & 43.23 & 22.47 & 6.88 & 67.00 & 40.63 \\

\midrule

\multicolumn{13}{c}{\textbf{\textit{Our Models}}} \\
\midrule
Qwen3-14B
& 3.79 & 19.65 & 41.40 & 32.74 & 25.32 & 38.82 & 0.83 & 31.00 & 1.12 & 7.16 & 29.00 & 8.62 \\

\hspace{0.4em}- SFT \textit{(w/o Mid-training)}
& 22.99 & 32.84 & 43.03 & 48.52 & 44.41 & 74.99 & 12.50 & 40.39 & 15.73 & 11.94 & 56.60 & 37.62 \\

\rowcolor{TrajRetrieval!12}
\hspace{0.4em}- \OurMODEL\ \textit{(Retrieval)}
& 23.93 & 35.95 & 47.55 & 57.01 & 42.07 & 76.75 & 23.61 & 42.84 & 22.47 & 29.82 & 66.40 & 46.24 \\

\rowcolor{TrajMultidocQA!12}
\hspace{0.4em}- \OurMODEL\ \textit{(MultidocQA)}
& 25.28 & 32.91 & 47.41 & 57.34 & 41.89 & 77.46 & 16.67 & 42.75 & 19.10 & 26.91 & 66.00 & 48.29 \\

\rowcolor{TrajStateTracking!12}
\hspace{0.4em}- \OurMODEL\ \textit{(State Tracking)}
& 22.91 & 37.13 & 50.56 & 56.09 & 45.86 & 77.19 & 29.17 & 42.06 & 26.97 & 29.28 & 67.20 & 48.97 \\

\rowcolor{TrajAll!16}
\textbf{\OurMODEL-14B-All}
& 22.83 & 35.26 & 47.85 & 57.72 & 41.35 & 75.62 & 20.83 & 42.03 & 24.72 & 26.08 & 67.60 & 49.66 \\

\midrule
Qwen3-30B-A3B
& 4.42 & 16.16 & 34.46 & 26.24 & 24.68 & 50.75 & 2.78 & 33.12 & 4.49 & 13.23 & 35.00 & 15.73 \\

\hspace{0.4em}- SFT \textit{(w/o Mid-training)}
& 19.91 & 31.90 & 38.07 & 51.99 & 44.23 & 72.01 & 12.50 & 38.20 & 5.62 & 18.29 & 51.40 & 37.76 \\

\rowcolor{TrajAll!16}
\textbf{\OurMODEL-30B-A3B-All}
& 20.85 & 36.57 & 46.12 & 56.29 & 42.88 & 72.24 & 18.06 & 38.60 & 21.35 & 26.49 & 64.80 & 47.33 \\

\bottomrule
\end{tabular}
}

\caption{
Overall SFT performance (BoN@3, \%) across diverse search, tool-use, and coding agentic benchmarks. BC-ZH, FSC-V2, and TB 2.1 denote BrowseComp-ZH, FinSearchComp V2, and Terminal-Bench 2.1, respectively. MCP-Atlas reports unthresholded claim coverage; scoring and three-trial aggregation are defined in Appendices~\ref{app:agent_evaluation} and~\ref{app:sft_evaluation_protocols}. Empty cells indicate unavailable or invalid evaluation results.
}
\label{tab:agent_results_pass}
\vspace{-0.25cm}
\end{table*}

\FloatBarrier
\subsection{General ICL Evaluation after Mid-training}
\label{app:general_icl}

We evaluate general capabilities before SFT, comparing Qwen3-14B-Base at
initialization, Combined Trajectories, and four \OurMODEL{} variants:
Retrieval, MultidocQA, State Tracking, and All. We report each trained model's
final checkpoint on ten selected evaluations under their respective ICL settings.

LiveCodeBench (LCB)~\citep{DBLP:conf/iclr/JainHGLYZWSSS25} tests code generation on
programming-contest problems, while the English and Chinese versions of
FullStackBench (FSB-EN and FSB-ZH)~\citep{cheng2024fullstack} cover practical
programming tasks across application domains and programming languages.
MMLU~\citep{hendrycks2020measuring} assesses knowledge and problem solving
across 57 subjects; MMLU-Redux~\citep{DBLP:conf/naacl/GemaLHDMSHZDMBMHKKM25} improves annotation quality
through manual review, and MMLU-Pro~\citep{wang2024mmlu} introduces harder,
reasoning-focused questions with up to ten answer options.
AGIEval-CN~\citep{zhong2024agieval} evaluates understanding and reasoning on
Chinese examination questions. KOR-Bench~\citep{ma2025kor} tests reasoning
with newly specified rules while reducing reliance on prior domain knowledge,
and PIQA~\citep{DBLP:conf/aaai/BiskZLGC20} assesses physical commonsense in everyday tasks.
Finally, the public MMMLU-Lite subset\footnote{\url{https://huggingface.co/datasets/opencompass/mmmlu_lite}}
evaluates multilingual knowledge understanding using translated MMLU questions.

\begin{figure}[!htbp]
    \centering
    \includegraphics[height=0.45\textheight]{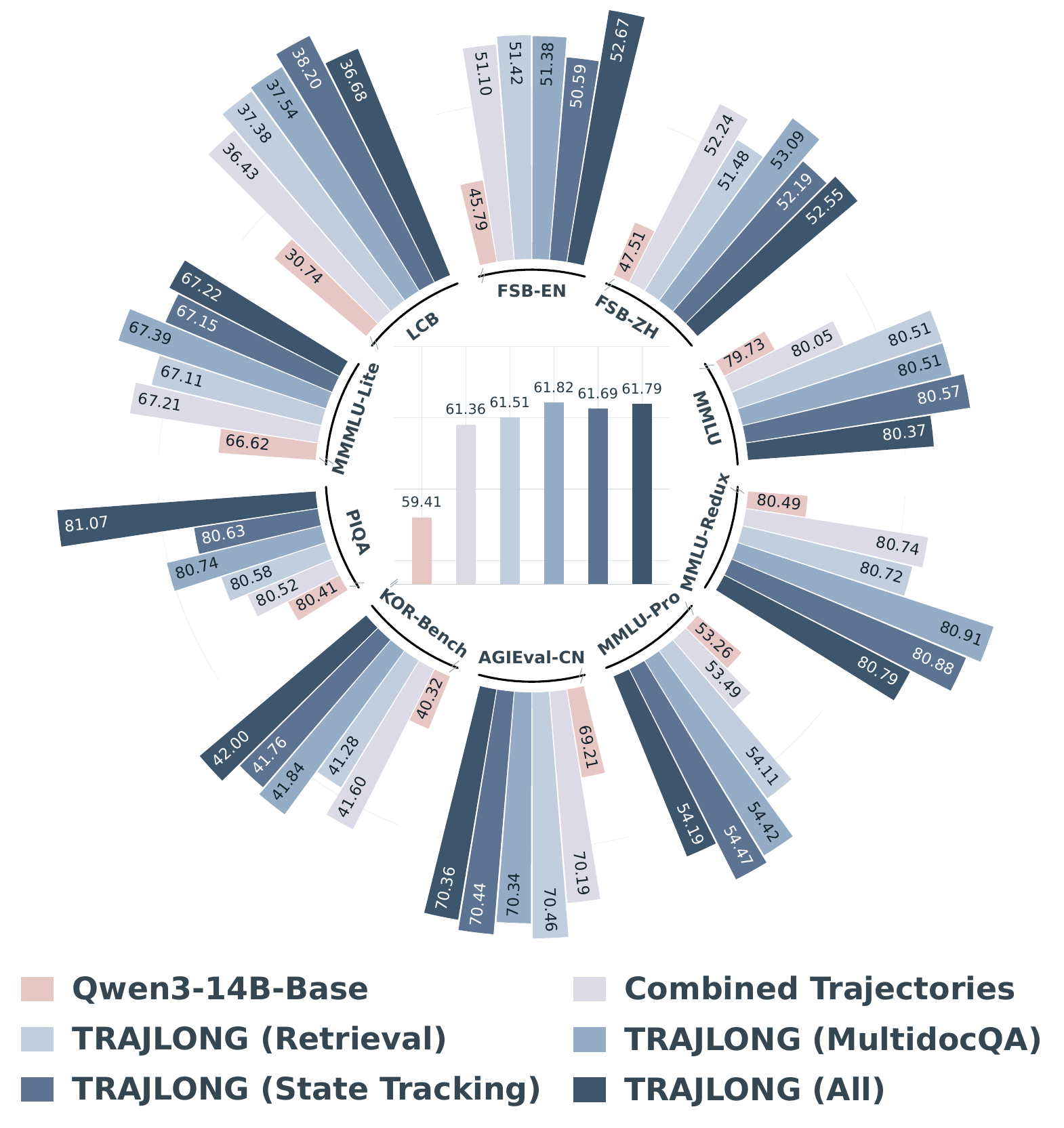}
    \caption{General ICL performance after mid-training on ten evaluations. Outer panels show task scores on individually zoomed axes; the center shows their unweighted mean.}
    \label{fig:general_icl_midtraining}
\end{figure}
\FloatBarrier

\subsection{Training Variants Produce Distinct Gain Profiles}
\label{app:icl_gain_profiles}

Figure~\ref{fig:icl_capability_gains} shows distinct patterns of joint improvement over Raw Trajectories. \textit{MultidocQA} leads on BrowseCompLong, RULER-QA1 ($+13.00$ pp), and Code Summary ($+4.25$ pp), and is the only compiled variant to improve Multi-hop Replanning. Gains also extend to Planning Preference and Tool Selection. These co-occurring benefits are consistent with shared demands for integrating contextual information in question answering, code understanding, and planning.

\textit{State Tracking} instead achieves the largest gain on Bug Localization ($+2.86$ pp) and ties \textit{MultidocQA} on Tool Selection, while declining on both planning proxies. The two variants thus share benefits for code understanding and tool choice but differ on fault localization and planning. \textit{State Tracking}'s leading gain on RULER-CWE, an aggregation task, also suggests overlap among the targeted long-context capabilities.

\textit{Retrieval} improves MRCR but regresses on Code Summary and Multi-hop Replanning: retrieval-oriented long-context gains need not accompany stronger agent reasoning. \textit{All} leads on GraphWalks, LongBench V2, and Planning Preference, yet does not preserve the RULER-CWE and Tool Selection gains of the individual variants. The mixture therefore combines strengths on selected tasks without retaining every specialized benefit. These findings suggest that trajectory organization shapes overlapping but distinct agent atomic capabilities.

\begingroup
\setlength{\intextsep}{5pt}
\begin{figure*}[!htbp]
    \centering
    \includegraphics[width=\textwidth,trim=0 24bp 0 8bp,clip]{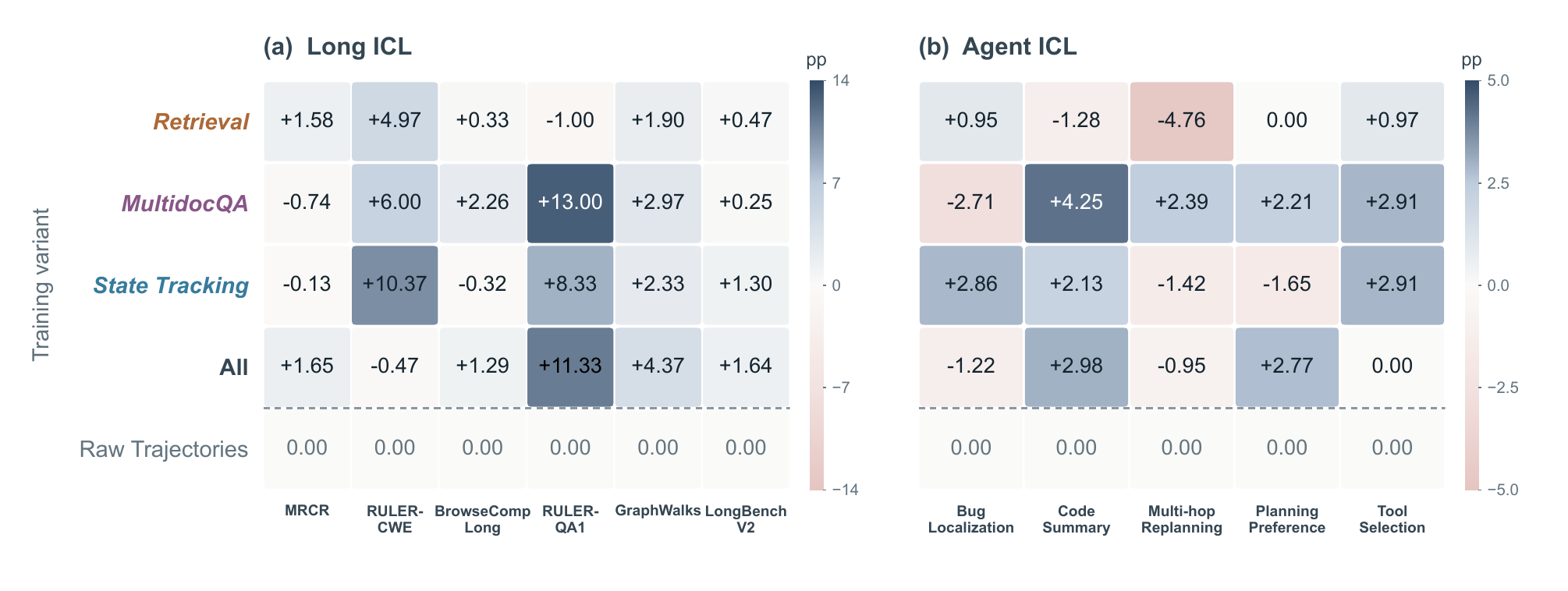}
    \caption{Long-context and Agent ICL accuracy gains over Raw Trajectories (pp). Dashed lines separate the Raw reference row.}
    \label{fig:icl_capability_gains}
\end{figure*}
\endgroup
\FloatBarrier

\subsection{ICL Proxies and Downstream Agent Performance}
\label{app:icl_to_sft_associations}

\begin{figure*}[htbp]
    \centering
    \includegraphics[width=\textwidth]{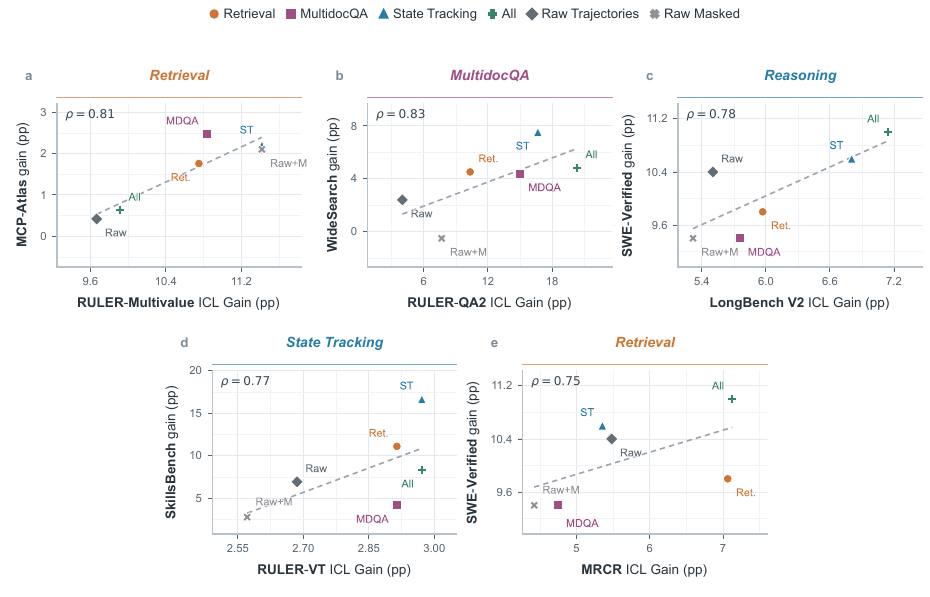}
    \caption{Correlations between long-context ICL gains and downstream agent gains after SFT. Each of the five panels contains six matched training variants. Axes show gains over Qwen3-14B-Base and SFT without mid-training, respectively (pp). Dashed lines are descriptive linear fits; $\rho$ denotes Spearman correlation.}
    \label{fig:icl_to_sft_association}
\end{figure*}

Figure~\ref{fig:icl_to_sft_association} compares final ICL proxy gains with downstream SFT Agent gains across matched training variants. RULER-Multivalue gains are positively associated with MCP-Atlas gains ($\rho=0.81$); RULER-QA2 and LongBench V2 provide additional comparisons for multi-document QA and reasoning. Panels~(d) and~(e) show RULER-VT--SkillsBench and MRCR--SWE-Verified, respectively. The RULER-MultiKey--BrowseComp-ZH comparison is shown separately in Figure~\ref{fig:icl_to_sft_retrieval_intro}.

For the additional pairs in Figure~\ref{fig:icl_to_sft_association}(d)--(e), SkillsBench and SWE-Verified use BoN@3 scores. RULER-VT averages six lengths (4K--128K); MRCR averages the 2-, 4-, and 8-needle settings at 256K. RULER-MultiKey in Figure~\ref{fig:icl_to_sft_retrieval_intro} averages six lengths (4K--128K).

\FloatBarrier
\subsection{Post-SFT Long-context--Agent Associations}
\label{app:sft_long_context_agent_associations}

\begin{figure}[!ht]
    \centering
    \includegraphics[width=\textwidth]{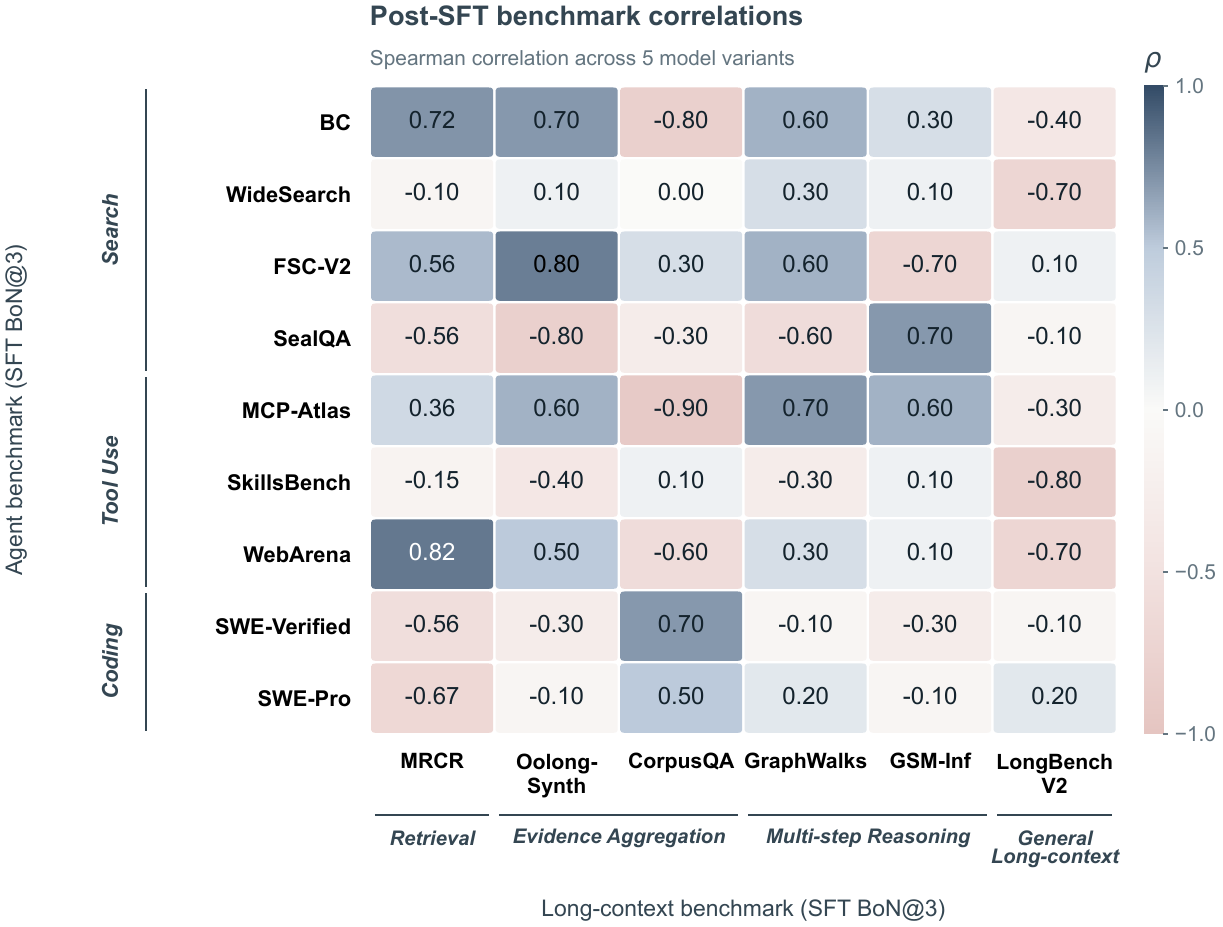}
    \caption{Long-to-Agent correlations after SFT (BoN@3) across five Qwen3-14B variants: Retrieval, MultidocQA, State Tracking, All, and Raw Trajectories. Cells show Spearman $\rho$ with average ranks for ties on a shared $[-1,1]$ scale.}
    \label{fig:sft_long_agent_bon_acc}
\end{figure}

\paragraph{Different Long-Context Capabilities Align with Different Agent Tasks.}
Figure~\ref{fig:sft_long_agent_bon_acc} shows task-dependent associations after SFT. Retrieval-oriented MRCR correlates positively with BrowseComp and FSC-V2 ($\rho=0.72$ and $0.56$). Evidence aggregation exhibits a different profile: CorpusQA correlates with SWE-Verified and SWE-Pro ($0.70$ and $0.50$), whereas MRCR is negatively correlated with both coding benchmarks. GraphWalks correlates with MCP-Atlas ($0.70$), but its association with WebArena is weaker ($0.30$). These complementary associations suggest that different long-context capabilities may be relevant to different downstream task demands. The correspondence is not uniform even within an agent category: MRCR correlates negatively with SealQA ($-0.56$), and SkillsBench shows no strong positive association with the plotted long-context benchmarks.

\FloatBarrier

\section{\OurMODEL{} Data Examples}
\label{app:trajlong_data_examples}

We present examples of the three \OurMODEL{} data types in
Tables~\ref{tab:case_retrieval}--\ref{tab:case_state_tracking}.
Long contexts are excerpted, with omissions marked as \texttt{...}.
The Retrieval example retains the search turns needed to answer its three
questions, with complete target passages and excerpted solutions.
State Tracking retains the key reads and all edits to the queried files;
MultidocQA retains selected evidence excerpts. Both use concise questions and
shortened or excerpted solutions while preserving the final answers. Identifying details
are normalized, and selected prompts and explanations are edited for readability.
Non-English text is translated consistently across the context, questions, and
answers. Retrieval questions continue the trajectory, whereas State Tracking
and MultidocQA questions follow the trajectory or evidence collection.

\subsection{Retrieval}
This example appends synthetic retrieval turns to an SEO agent trajectory for
mid-training. The model locates earlier search results by topic and rank, then
reproduces them verbatim with any requested prefix. Thoughts and answers are
excerpted for display.

\begingroup
\small
\definecolor{RetrievalUser}{HTML}{285C91}
\definecolor{RetrievalAssistant}{HTML}{914C65}
\definecolor{RetrievalObservation}{HTML}{806A49}
\definecolor{RetrievalFocus}{HTML}{5B467A}
\setlength{\tabcolsep}{4pt}
\begin{longtable}{>{\raggedright\arraybackslash}p{0.16\linewidth}>{\raggedright\arraybackslash}p{0.80\linewidth}}
\toprule
\multicolumn{2}{l}{\cellcolor{TrajRetrieval!12}\textbf{\OurMODEL{} / Retrieval}} \\
\midrule
\endfirsthead
\toprule
\multicolumn{2}{l}{\cellcolor{TrajRetrieval!12}\textbf{\OurMODEL{} / Retrieval (continued)}} \\
\midrule
\endhead
\midrule
\multicolumn{2}{r}{\textit{Continued on next page}} \\
\endfoot
\bottomrule
\caption{A \OurMODEL{} (Retrieval) example.\label{tab:case_retrieval}} \\
\endlastfoot

\rowcolor{RetrievalUser!7}
\textcolor{RetrievalUser}{\textbf{User}} & Run today's SEO workflow for example-store.example. Review the saved history and strategy, rotate the target markets across India, Sri Lanka, and Nepal, and avoid duplicate publications. \\
\multicolumn{2}{c}{\textcolor{gray}{\textit{... [1 assistant turn omitted; $\approx$2.9k tokens] ...}}} \\
\rowcolor{RetrievalAssistant!7}
\textcolor{RetrievalAssistant}{\textbf{Assistant}} & \textbf{Answer:} Today is Monday, so I will check the site's indexing status and recent articles, then prepare the weekly SEO report. \\*
\rowcolor{RetrievalAssistant!7}
 & \textbf{Tool Call:} web\_search(\{ "query": "site:example-store.example", "max\_results": 10 \}) \\
\rowcolor{RetrievalObservation!6}
\textcolor{RetrievalObservation}{\textbf{Observation}} & {[example-store.example (search results)]} \\*
\rowcolor{RetrievalObservation!6}
 & \textbf{Result 1:} finding your goldmine: the ultimate guide to successful dollar store site selection ... \\
\rowcolor{RetrievalObservation!6}
 & \textbf{Results 2–3:} ... \\
\rowcolor{RetrievalObservation!6}
 & \textbf{Result 4:} 30\% off for 3+ items! christmas sale! ... \\
\rowcolor{RetrievalObservation!6}
 & \textbf{Result 5:} 30\% off for 3+ items! christmas sale! [EMAIL] the ExampleCos collection inspired by the movie \textbf{poor things} features a range of stylish clothing and accessories that capture the unique aesthetic of the film. this collection includes vibrant costumes that reflect the whimsical and eclectic fashion seen in the movie, showcasing bold colors and interesting patterns. the clothing is crafted from high-quality materials, ensuring comfort and durability, perfect for cosplay or themed events. \\
\multicolumn{2}{c}{\textcolor{gray}{\textit{... [2 assistant turns omitted; $\approx$2.4k tokens] ...}}} \\
\rowcolor{RetrievalAssistant!7}
\textcolor{RetrievalAssistant}{\textbf{Assistant}} & \textbf{Tool Call:} web\_search(\{ "query": "site:example-store.example Sri Lanka product mix optimization", "max\_results": 5 \}) \\
\rowcolor{RetrievalObservation!6}
\textcolor{RetrievalObservation}{\textbf{Observation}} & {[Sri Lanka product-mix optimization (search results)]} \\*
\rowcolor{RetrievalObservation!6}
 & \textbf{Result 1:} This paper propose a practical optimization problem of product-mix based on labor capacity, machine capacity, raw materials and demand constrains in a \textbf{Sri Lankan tea producing company}. The objective is to profit maximization, satisfying all \\
\rowcolor{RetrievalObservation!6}
 & \textbf{Results 2–5:} ... \\
\multicolumn{2}{c}{\textcolor{gray}{\textit{... [6 assistant turns omitted; $\approx$3.6k tokens] ...}}} \\
\rowcolor{RetrievalAssistant!7}
\textcolor{RetrievalAssistant}{\textbf{Assistant}} & \textbf{Tool Call:} web\_search(\{ "query": "site:example-store.example India inventory turnover", "max\_results": 5 \}) \\
\rowcolor{RetrievalObservation!6}
\textcolor{RetrievalObservation}{\textbf{Observation}} & {[India inventory turnover (search results)]} \\*
\rowcolor{RetrievalObservation!6}
 & \textbf{Result 1:} PDF | On Sep 18, 2021, Aarti Patki published Analysis of Inventory Turnover Ratio and Its Impact on Profitability of Enterprises in Indian Retail Industry Author ... \\
\rowcolor{RetrievalObservation!6}
 & \textbf{Results 2–4:} ... \\
\rowcolor{RetrievalObservation!6}
 & \textbf{Result 5:} December 30, 2025 · \textbf{Analysis of inventory turnover performance within Indian supermarket} Ashish Kumar Bhateja, Abhishek Kumar Shukla Mechanical Engineering, Gulzar Group of Institutions, \mbox{Khanna …} \\
\multicolumn{2}{c}{\textcolor{gray}{\textit{... [4 assistant turns omitted; $\approx$3.4k tokens] ...}}} \\
\multicolumn{2}{l}{\textcolor{RetrievalFocus}{(Synthetic retrieval turns for mid-training)}} \\*
\addlinespace[0.3em]
\rowcolor{RetrievalUser!7}
\textcolor{RetrievalUser}{\textbf{User}} & \textbf{From the "example-store.example" search, reproduce result 5 (1-indexed) verbatim.} \\*
\rowcolor{RetrievalAssistant!7}
\textcolor{RetrievalAssistant}{\textbf{Assistant}} & \textbf{Thought:} \textit{Results 4 and 5 share the same promotional opening. ... The requested fifth entry is the Poor Things collection passage. Copy it verbatim, without a result label or prefix.} \\*
\rowcolor{RetrievalAssistant!7}
 & \textbf{Answer:} 30\% off for 3+ items! christmas sale! [EMAIL] the ExampleCos collection inspired by the movie \textbf{poor things} [... omitted for display ...] perfect for cosplay or themed events. \\
\addlinespace[0.35em]
\rowcolor{RetrievalUser!7}
\textcolor{RetrievalUser}{\textbf{User}} & \textbf{From the "Sri Lanka product-mix optimization" search, reproduce result 1 (1-indexed) verbatim. Prepend 0qM5Sa8J51 followed by a space.} \\*
\rowcolor{RetrievalAssistant!7}
\textcolor{RetrievalAssistant}{\textbf{Assistant}} & \textbf{Thought:} \textit{The first result describes product-mix optimization for a Sri Lankan tea producer. ... Preserve its wording and incomplete ending, and prepend 0qM5Sa8J51 followed by one space.} \\*
\rowcolor{RetrievalAssistant!7}
 & \textbf{Answer:} 0qM5Sa8J51 This paper propose a practical optimization problem of product-mix [... omitted for display ...] in a \textbf{Sri Lankan tea producing company}. The objective is to profit maximization, satisfying all \\
\addlinespace[0.35em]
\rowcolor{RetrievalUser!7}
\textcolor{RetrievalUser}{\textbf{User}} & \textbf{From the "India inventory turnover" search, reproduce result 5 (1-indexed) verbatim. Prepend EVfcnq2nOO followed by a space.} \\*
\rowcolor{RetrievalAssistant!7}
\textcolor{RetrievalAssistant}{\textbf{Assistant}} & \textbf{Thought:} \textit{The fifth result is the supermarket study dated December 30, 2025. ... Include its date and trailing ellipsis, omit the search heading, and prepend EVfcnq2nOO followed by one space.} \\*
\rowcolor{RetrievalAssistant!7}
 & \textbf{Answer:} EVfcnq2nOO December 30, 2025 · \textbf{Analysis of inventory turnover performance within Indian supermarket} Ashish Kumar Bhateja, Abhishek Kumar Shukla [... omitted for display ...] \mbox{Khanna …} \\
\end{longtable}
\endgroup

\subsection{MultidocQA}
This example assembles code evidence from an agent trajectory into a collection
with evidence IDs and versions. The questions combine evidence across files
and versions to identify login endpoints, infer role-dependent navigation,
and compare notification-filtering behavior. The original context contains
111 evidence items and approximately 107K tokens; this compact display retains
the eight items used by the three questions below.

\begingroup
\small
\definecolor{MultiQAFocus}{HTML}{5B467A}
\definecolor{MultiQAEvidence}{HTML}{285C91}
\begin{longtable}{>{\raggedright\arraybackslash}p{0.95\linewidth}}
\toprule
\cellcolor{TrajMultidocQA!12}\textbf{\OurMODEL{} / MultidocQA} \\
\midrule
\endfirsthead
\toprule
\cellcolor{TrajMultidocQA!12}\textbf{\OurMODEL{} / MultidocQA (continued)} \\
\midrule
\endhead
\midrule
\multicolumn{1}{r}{\textit{Continued on next page}} \\
\endfoot
\bottomrule
\caption{A \OurMODEL{} (MultidocQA) example.\label{tab:case_multidocqa}} \\
\endlastfoot

\textcolor{MultiQAFocus}{You are given a set of evidence items below, each marked with a unique ID,
together with its title and version when available. Read the evidence
carefully, identify the items relevant to each question, and answer the
question or questions that follow.
Evidence IDs are provided to help locate and distinguish evidence items.
Use them to keep track of the relevant evidence while solving each question.
Follow each question's own output requirements.} \\

\textcolor{MultiQAFocus}{\textit{... [103 additional evidence items omitted from this compact display] ...}} \\
\cellcolor{MultiQAEvidence!5}\textcolor{MultiQAEvidence}{\textbf{[EVIDENCE id=file-d4ba5ce340e9 | title=AuthRepository.kt | version=1]}}\newline
... class AuthRepository(...) \{ ...\newline
private val api: AuthApi by lazy \{ ApiClient.createApi(AuthApi::class.java) \}\newline
suspend fun loginWithPassword(...): Result<LoginResponse> = safeCall \{\newline
\hspace*{0.7em}val response = api.loginWithPassword(LoginRequest(phone, password)); ... \}\newline
suspend fun loginWithSms(...): Result<LoginResponse> = safeCall \{\newline
\hspace*{0.7em}val response = api.loginWithSms(SmsLoginRequest(phone, code)); ... \}\newline
... \} \\
\addlinespace[0.3em]
\cellcolor{MultiQAEvidence!5}\textcolor{MultiQAEvidence}{\textbf{[EVIDENCE id=file-ce5214a29b14 | title=AuthApi.kt | version=1]}}\newline
... interface AuthApi \{\newline
@POST("/api/v1/auth/tenant/login") suspend fun loginWithPassword(...) ...\newline
@POST("/api/v1/auth/sms/send") suspend fun sendSmsCode(...) ...\newline
@POST("/api/v1/auth/sms/login") suspend fun loginWithSms(...) ...\newline
... \} \\
\addlinespace[0.3em]
\cellcolor{MultiQAEvidence!5}\textcolor{MultiQAEvidence}{\textbf{[EVIDENCE id=file-f879713c7014 | title=AppNavGraph.kt | version=1]}}\newline
... LoginScreen(onLoginSuccess = \{\newline
val user = authState.user\newline
val homeRoute = when \{ user?.isAdmin() == true -> Screen.AdminHome.route\newline
\hspace*{0.7em}else -> Screen.WorkerHome.route \}\newline
navController.navigate(homeRoute) \{ popUpTo(Screen.Login.route) \{ inclusive = true \} \}\newline
\}) ... \\
\addlinespace[0.3em]
\cellcolor{MultiQAEvidence!5}\textcolor{MultiQAEvidence}{\textbf{[EVIDENCE id=file-8c0dc44e8e97 | title=NotificationApi.kt | version=1]}}\newline
... @GET("/api/v1/notifications") suspend fun getNotifications(\newline
@Query("page") page: Int = 1, @Query("limit") limit: Int = 20,\newline
@Query("is\_read") isRead: Boolean? = null): PaginatedResponse<NotificationResponse> ... \\
\addlinespace[0.3em]
\cellcolor{MultiQAEvidence!5}\textcolor{MultiQAEvidence}{\textbf{[EVIDENCE id=file-043ad1aa30d9 | title=NotificationApi.kt | version=2]}}\newline
... @GET("/api/v1/notifications") suspend fun getNotifications(\newline
@retrofit2.http.Query("page") page: Int = 1, @retrofit2.http.Query("limit") limit: Int = 20\newline
): ApiResponse<List<Notification>> ... \\
\addlinespace[0.3em]
\cellcolor{MultiQAEvidence!5}\textcolor{MultiQAEvidence}{\textbf{[EVIDENCE id=file-302481661e72 | title=NotificationApi.kt | version=3]}}\newline
... @GET("/api/v1/notifications") suspend fun getNotifications(\newline
@Query("page") page: Int = 1, @Query("limit") limit: Int = 20,\newline
@Query("is\_read") isRead: String? = null): PaginatedResponse<Notification> ... \\
\addlinespace[0.3em]
\cellcolor{MultiQAEvidence!5}\textcolor{MultiQAEvidence}{\textbf{[EVIDENCE id=file-fd9d2fee6a0c | title=notificationController.js | version=1]}}\newline
... async function listNotifications(req, res) \{\newline
const \{ type, is\_read, page = 1, limit = 20 \} = req.query\newline
const \{ user\_id, user\_type \} = req.user; const where = \{ user\_id, user\_type \}\newline
if (type) where.type = type\newline
if (is\_read !== undefined \&\& is\_read !== '') \{\newline
\hspace*{0.7em}where.is\_read = is\_read === '1' || is\_read === 'true'\newline
\}\newline
... Notification.findAndCountAll(\{ where, ... \}) ... \} \\
\addlinespace[0.3em]
\cellcolor{MultiQAEvidence!5}\textcolor{MultiQAEvidence}{\textbf{[EVIDENCE id=file-4150517855d6 | title=User.kt | version=1]}}\newline
... data class User(... val role: String, ...) \{\newline
fun isAdmin(): Boolean = role == "admin"; ... \} \\

\addlinespace[0.6em]
\cellcolor{MultiQAFocus!6}\textcolor{MultiQAFocus}{\textbf{Question 1.}}
Which POST endpoints does AuthRepository use for password and SMS login?
Return the two paths as a JSON array in alphabetical order. \\*
\cellcolor{CaseAnswer!6}\textcolor{CaseAnswer}{\textbf{Reason:}}
\textit{The two repository methods must first be mapped to the corresponding
Retrofit methods. In file-d4ba5ce340e9, loginWithPassword calls
api.loginWithPassword(LoginRequest(phone, password)), while loginWithSms calls
api.loginWithSms(SmsLoginRequest(phone, code)). ...
In file-ce5214a29b14, the @POST annotation on AuthApi.loginWithPassword
specifies "/api/v1/auth/tenant/login". The annotation on AuthApi.loginWithSms
specifies "/api/v1/auth/sms/login". The same interface also declares
"/api/v1/auth/sms/send", but that endpoint belongs to sendSmsCode and
\textbf{only sends a verification code}; it is not the SMS login endpoint
requested by the question. ... Their common prefix is "/api/v1/auth/",
so alphabetical ordering places "sms" before "tenant". ...} \\*
\cellcolor{CaseAnswer!6}\textcolor{CaseAnswer}{\textbf{Answer:}}
{["/api/v1/auth/sms/login", "/api/v1/auth/tenant/login"]} \\
\addlinespace[0.6em]
\cellcolor{MultiQAFocus!6}\textcolor{MultiQAFocus}{\textbf{Question 2.}}
After a user with role 'designer' logs in through LoginScreen, which route is selected?
Give the popUpTo target and its inclusive setting. \\*
\cellcolor{CaseAnswer!6}\textcolor{CaseAnswer}{\textbf{Reason:}}
\textit{The role check and the navigation callback are defined in separate
evidence items. In file-4150517855d6, User.isAdmin() returns role == "admin".
A user whose role is "designer" therefore does not satisfy the administrator
check. ... In file-f879713c7014, the LoginScreen onLoginSuccess callback reads
authState.user and chooses homeRoute with a when expression. It selects
Screen.AdminHome.route only when user?.isAdmin() == true; otherwise, it
selects Screen.WorkerHome.route. \textbf{The designer user follows this second
branch.} ... The callback then calls navController.navigate(homeRoute) with
popUpTo(Screen.Login.route) \{ inclusive = true \}. Thus the navigation pops
the back stack up to and including the login destination, removing the login
screen. ...} \\*
\cellcolor{CaseAnswer!6}\textcolor{CaseAnswer}{\textbf{Answer:}}
Screen.WorkerHome.route; popUpTo(Screen.Login.route), inclusive = true. \\
\addlinespace[0.6em]
\cellcolor{MultiQAFocus!6}\textcolor{MultiQAFocus}{\textbf{Question 3.}}
For GET /api/v1/notifications, which statements about the Android API versions
and backend is\_read handling are correct? Select all that apply;
return comma-separated letters in alphabetical order. \\*
\cellcolor{MultiQAFocus!6}
(A) Android v2 declares isRead: Int? = 0 as a query parameter.\newline
(B) Android v1 declares isRead: Boolean? = null as a query parameter.\newline
(C) Android v3 declares isRead: String? = null as a query parameter.\newline
(D) Omitting is\_read makes the backend return only unread notifications.\newline
(E) Android v2 omits the is\_read query parameter from getNotifications.\newline
(F) The backend applies an is\_read filter only when the parameter is neither
undefined nor empty, interpreting the strings '1' and 'true' as true. \\*
\cellcolor{CaseAnswer!6}\textcolor{CaseAnswer}{\textbf{Reason:}}
\textit{The three Android evidence items must be checked separately because
they contain different versions of the same interface. In version 1
(file-8c0dc44e8e97), getNotifications declares @Query("is\_read") isRead:
Boolean? = null, making (B) correct. Version 2 (file-043ad1aa30d9) declares
only page and limit in that method, so (E) is correct and (A) is false. ...
Version 3 (file-302481661e72) declares @Query("is\_read") isRead: String? = null,
making (C) correct. The backend logic is in listNotifications in
file-fd9d2fee6a0c. ... It adds where.is\_read only when is\_read !== undefined
\&\& is\_read !== ''. Inside that condition, it evaluates is\_read === '1'
|| is\_read === 'true'. This directly supports (F). When the request omits
is\_read, the condition is false and listNotifications adds
\textbf{no read-status filter}. Omission therefore does not mean "unread only",
so (D) is false. ...} \\*
\cellcolor{CaseAnswer!6}\textcolor{CaseAnswer}{\textbf{Answer:}} B,C,E,F \\
\end{longtable}
\endgroup

\clearpage
\subsection{State Tracking}
This example follows a coding agent adding font-size controls. The questions
trace file states across reads and edits, distinguishing initial content from
later changes and identifying the last operation that wrote a target fragment.
All successful Edit and Write tool calls on the queried files are retained;
omitted turns concern other files or actions unrelated to the answers. Tool-call
arguments marked as abridged are shortened only for display.

\begingroup
\small
\definecolor{StateUser}{HTML}{285C91}
\definecolor{StateAssistant}{HTML}{914C65}
\definecolor{StateObservation}{HTML}{806A49}
\definecolor{StateSystem}{HTML}{555555}
\definecolor{StateFocus}{HTML}{5B467A}
\setlength{\tabcolsep}{4pt}
\begin{longtable}{>{\raggedright\arraybackslash}p{0.16\linewidth}>{\raggedright\arraybackslash}p{0.80\linewidth}}
\toprule
\multicolumn{2}{l}{\cellcolor{TrajStateTracking!12}\textbf{\OurMODEL{} / State Tracking}} \\
\midrule
\endfirsthead
\toprule
\multicolumn{2}{l}{\cellcolor{TrajStateTracking!12}\textbf{\OurMODEL{} / State Tracking (continued)}} \\
\midrule
\endhead
\midrule
\multicolumn{2}{r}{\textit{Continued on next page}} \\
\endfoot
\bottomrule
\caption{A \OurMODEL{} (State Tracking) example.\label{tab:case_state_tracking}} \\
\endlastfoot

\multicolumn{2}{>{\raggedright\arraybackslash}p{0.96\linewidth}}{\textcolor{StateFocus}{The following is a historical agent trajectory enclosed between [BEGIN\_OF\_TRAJECTORY] and [END\_OF\_TRAJECTORY].}} \\
\addlinespace[0.2em]
\multicolumn{2}{l}{\textcolor{StateFocus}{[BEGIN\_OF\_TRAJECTORY]}} \\*
\rowcolor{StateSystem!7}
\textcolor{StateSystem}{\textbf{System}} & You are an AI coding assistant. \\
\multicolumn{2}{c}{\textcolor{gray}{\textit{... [20 assistant turns omitted; $\approx$28.0k tokens] ...}}} \\
\rowcolor{StateUser!7}
\textcolor{StateUser}{\textbf{User}} & Can we adjust the font size? It looks a little too small. \\*
\rowcolor{StateAssistant!7}
\textcolor{StateAssistant}{\textbf{Assistant}} & \textbf{Tool Call:} Read(\{"file\_path": ".../src/renderer/index.css"\}) \\*
\rowcolor{StateObservation!6}
\textcolor{StateObservation}{\textbf{Observation}} & ... \\*
\rowcolor{StateObservation!6}
 & 10→ :root \{ \\*
\rowcolor{StateObservation!6}
 & 11→ \hspace*{0.7em}/* Cyberpunk Color Tokens */ \\*
\rowcolor{StateObservation!6}
 & 12→ \hspace*{0.7em}--background: \#0a0a0f; \\*
\rowcolor{StateObservation!6}
 & 13→ \hspace*{0.7em}--foreground: \#e0e0e0; \\*
\rowcolor{StateObservation!6}
 & ... \\
\multicolumn{2}{c}{\textcolor{gray}{\textit{... [3 assistant turns omitted; $\approx$9.8k tokens] ...}}} \\
\rowcolor{StateAssistant!7}
\textcolor{StateAssistant}{\textbf{Assistant}} & \textbf{Answer:} Add separate UI and terminal font-size variables. \\*
\rowcolor{StateAssistant!7}
 & \textbf{Tool Call (abridged):} Edit(\{"file\_path": ".../src/renderer/index.css", "old\_string": ":root \{ /* Cyberpunk Color Tokens */", "new\_string": ":root \{ /* Font Size Tokens (adjustable via UI) */ --ui-font-size: 12px; --terminal-font-size: 12px; /* Cyberpunk Color Tokens */"\}) \\*
\rowcolor{StateObservation!6}
\textcolor{StateObservation}{\textbf{Observation}} & Updated index.css successfully. \\
\addlinespace[0.25em]
\rowcolor{StateAssistant!7}
\textcolor{StateAssistant}{\textbf{Assistant}} & \textbf{Answer:} Apply the UI font-size variable to the body rule. \\*
\rowcolor{StateAssistant!7}
 & \textbf{Tool Call (abridged):} Edit(\{"file\_path": ".../src/renderer/index.css", "old\_string": "body \{ font-family: ...; ... \}", "new\_string": "body \{ font-family: ...; font-size: var(--ui-font-size); ... \}"\}) \\*
\rowcolor{StateObservation!6}
\textcolor{StateObservation}{\textbf{Observation}} & Updated index.css successfully. \\
\multicolumn{2}{c}{\textcolor{gray}{\textit{... [3 assistant turns omitted; $\approx$1.4k tokens] ...}}} \\
\rowcolor{StateAssistant!7}
\textcolor{StateAssistant}{\textbf{Assistant}} & \textbf{Answer:} Inspect TerminalGrid before updating terminal font sizes. \\*
\rowcolor{StateAssistant!7}
 & \textbf{Tool Call:} Read(\{"file\_path": ".../src/renderer/components/TerminalGrid.jsx"\}) \\*
\rowcolor{StateObservation!6}
\textcolor{StateObservation}{\textbf{Observation}} & ... \\*
\rowcolor{StateObservation!6}
 & 126→ \hspace*{1.4em}setTerminalOrder((prev) => \{ \\*
\rowcolor{StateObservation!6}
 & 127→ \hspace*{2.1em}let rows = prev.map((row) => row.filter((id) => id !== agentId)); \\*
\rowcolor{StateObservation!6}
 & 128→ \hspace*{2.1em}rows[targetRowIdx] = [ \\*
\rowcolor{StateObservation!6}
 & 129→ \hspace*{2.8em}...rows[targetRowIdx].slice(0, targetColIdx), \\*
\rowcolor{StateObservation!6}
 & 130→ \hspace*{2.8em}agentId, \\*
\rowcolor{StateObservation!6}
 & 131→ \hspace*{2.8em}...rows[targetRowIdx].slice(targetColIdx), \\*
\rowcolor{StateObservation!6}
 & 132→ \hspace*{2.1em}]; \\*
\rowcolor{StateObservation!6}
 & 133→ \hspace*{2.1em}return rows.filter((r) => r.length > 0); \\*
\rowcolor{StateObservation!6}
 & 134→ \hspace*{1.4em}\}); \\*
\rowcolor{StateObservation!6}
 & ... \\
\multicolumn{2}{c}{\textcolor{gray}{\textit{... [4 assistant turns omitted; $\approx$2.2k tokens] ...}}} \\
\rowcolor{StateAssistant!7}
\textcolor{StateAssistant}{\textbf{Assistant}} & \textbf{Answer:} Expose updateAllFontSize to update existing terminals. \\*
\rowcolor{StateAssistant!7}
 & \textbf{Tool Call (abridged):} Edit(\{"file\_path": ".../src/renderer/components/TerminalGrid.jsx", "old\_string": "useImperativeHandle(ref, () => (\{ ... \}));", "new\_string": "useImperativeHandle(ref, () => (\{ ... updateAllFontSize: (size) => \{ Object.values(termRefs.current).forEach((t) => t?.updateFontSize?.(size)); \}, \}));"\}) \\*
\rowcolor{StateObservation!6}
\textcolor{StateObservation}{\textbf{Observation}} & Updated TerminalGrid.jsx successfully. \\
\multicolumn{2}{c}{\textcolor{gray}{\textit{... [33 assistant turns omitted; $\approx$27.8k tokens] ...}}} \\
\multicolumn{2}{l}{\textcolor{StateFocus}{[END\_OF\_TRAJECTORY]}} \\
\addlinespace[0.3em]
\multicolumn{2}{>{\raggedright\arraybackslash}p{0.96\linewidth}}{\textcolor{StateFocus}{Return only \{"operation":"...","successful\_index\_in\_file":N\}. Count only successful Edit, Write, and Delete tool calls, separately by operation type and file, starting at 1. Read and Bash tool calls (including git commands), failed calls, and operations on other files do not count. Use \textbf{anchor} with index 0 for a fragment present at the trajectory's start.}} \\*
\addlinespace[0.4em]
\multicolumn{2}{>{\raggedright\arraybackslash}p{0.96\linewidth}}{\cellcolor{StateFocus!6}\textcolor{StateFocus}{\textbf{Question 1:}} Identify the first appearance of this fragment in \textbf{TerminalGrid.jsx}. Return \textbf{anchor} if it was present at the start; otherwise, identify the successful edit or write that first introduced it.} \\*
\multicolumn{2}{>{\raggedright\arraybackslash}p{0.96\linewidth}}{\cellcolor{StateFocus!6}{[TARGET\_FRAGMENT]}} \\*
\multicolumn{2}{>{\raggedright\arraybackslash}p{0.96\linewidth}}{\cellcolor{StateFocus!6}let rows = prev.map((row) => row.filter((id) => id !== agentId));} \\*
\multicolumn{2}{>{\raggedright\arraybackslash}p{0.96\linewidth}}{\cellcolor{StateFocus!6}rows[targetRowIdx] = [} \\*
\multicolumn{2}{>{\raggedright\arraybackslash}p{0.96\linewidth}}{\cellcolor{StateFocus!6}{[END\_OF\_TARGET\_FRAGMENT]}} \\*
\multicolumn{2}{>{\raggedright\arraybackslash}p{0.96\linewidth}}{\cellcolor{CaseAnswer!6}\textcolor{CaseAnswer}{\textbf{Reason:}} \textit{The first read of TerminalGrid.jsx already shows the target at lines 127–128, before any successful edit or write to this file. Its only later edit adds updateAllFontSize to useImperativeHandle and leaves the target fragment unchanged. The read observes existing content rather than creating it. Therefore, the fragment belongs to the initial state: anchor, index 0.}} \\*
\multicolumn{2}{>{\raggedright\arraybackslash}p{0.96\linewidth}}{\cellcolor{CaseAnswer!6}\textcolor{CaseAnswer}{\textbf{Answer:}} \{"operation":"anchor","successful\_index\_in\_file":0\}} \\
\addlinespace[0.65em]
\multicolumn{2}{>{\raggedright\arraybackslash}p{0.96\linewidth}}{\cellcolor{StateFocus!6}\textcolor{StateFocus}{\textbf{Question 2:}} This fragment remains in \textbf{index.css} at the end of the trajectory. Which successful edit or write last wrote it?} \\*
\multicolumn{2}{>{\raggedright\arraybackslash}p{0.96\linewidth}}{\cellcolor{StateFocus!6}{[TARGET\_FRAGMENT]}} \\*
\multicolumn{2}{>{\raggedright\arraybackslash}p{0.96\linewidth}}{\cellcolor{StateFocus!6}/* Font Size Tokens (adjustable via UI) */} \\*
\multicolumn{2}{>{\raggedright\arraybackslash}p{0.96\linewidth}}{\cellcolor{StateFocus!6}--ui-font-size: 12px;} \\*
\multicolumn{2}{>{\raggedright\arraybackslash}p{0.96\linewidth}}{\cellcolor{StateFocus!6}--terminal-font-size: 12px;} \\*
\multicolumn{2}{>{\raggedright\arraybackslash}p{0.96\linewidth}}{\cellcolor{StateFocus!6}{[END\_OF\_TARGET\_FRAGMENT]}} \\*
\multicolumn{2}{>{\raggedright\arraybackslash}p{0.96\linewidth}}{\cellcolor{CaseAnswer!6}\textcolor{CaseAnswer}{\textbf{Reason:}} \textit{The initial read of index.css has no font-size token block. Successful edit 1 inserts the comment and both 12px variables under :root. Successful edit 2 only adds font-size: var(--ui-font-size); to the body rule, leaving those tokens intact. No later edit or write rewrites the target fragment. Thus, edit 1 is the last operation that wrote this fragment, even though edit 2 is the file's last edit.}} \\*
\multicolumn{2}{>{\raggedright\arraybackslash}p{0.96\linewidth}}{\cellcolor{CaseAnswer!6}\textcolor{CaseAnswer}{\textbf{Answer:}} \{"operation":"edit","successful\_index\_in\_file":1\}} \\
\end{longtable}
\endgroup

\end{document}